\documentclass[10pt,twocolumn,letterpaper]{article}

\usepackage{cvpr}
\usepackage{times}
\usepackage{epsfig}
\usepackage{graphicx}
\usepackage{amsmath}
\usepackage{amssymb}
\usepackage{booktabs}
\usepackage{tabularx}
\usepackage{subfigure}

\usepackage{amsfonts}
\newcommand{\ms}[1]{\tiny{$\pm$#1}}
\usepackage{color, colortbl}
\definecolor{Gray}{gray}{0.9}
\usepackage{flafter}

\makeatletter\renewcommand\paragraph{\@startsection{paragraph}{4}{\z@}
  {.5em \@plus1ex \@minus.2ex}{-.5em}{\normalfont\normalsize\bfseries}}\makeatother

\usepackage{hyperref}
\hypersetup{pagebackref=true,breaklinks=true,letterpaper=true,colorlinks,bookmarks=false}

\cvprfinalcopy % *** Uncomment this line for the final submission

\ifcvprfinal\fi
\begin{document}

%%%%%%%%% TITLE
\title{CReST: A Class-Rebalancing Self-Training Framework \\ for Imbalanced Semi-Supervised Learning}

\author{
Chen Wei\textsuperscript{1}\thanks{Work done while an intern at Google.}~~~
Kihyuk Sohn\textsuperscript{2}~~~
Clayton Mellina\textsuperscript{2}~~~
Alan Yuille\textsuperscript{1}~~~
Fan Yang\textsuperscript{2}\\
\textsuperscript{1}Johns Hopkins University~~~
\textsuperscript{2}Google Cloud AI
}

\maketitle
%\thispagestyle{empty}

%%%%%%%%% ABSTRACT
\begin{abstract}
Semi-supervised learning on class-imbalanced data, although a realistic problem, has been under studied. While existing semi-supervised learning (SSL) methods are known to perform poorly on minority classes, we find that they still generate high precision pseudo-labels on minority classes. By exploiting this property, in this work, we propose Class-Rebalancing Self-Training (CReST), a simple yet effective framework to improve existing SSL methods on class-imbalanced data. CReST iteratively retrains a baseline SSL model with a labeled set expanded by adding pseudo-labeled samples from an unlabeled set, where pseudo-labeled samples from minority classes are selected more frequently according to an estimated class distribution. We also propose a progressive distribution alignment to adaptively adjust the rebalancing strength dubbed CReST+. We show that CReST and CReST+ improve state-of-the-art SSL algorithms on various class-imbalanced datasets and consistently outperform other popular rebalancing methods. Code has been made available at \url{https://github.com/google-research/crest}.

\end{abstract}

%%%%%%%%% BODY TEXT

\vspace{-0.25cm}
\section{Introduction}
\vspace{-0.1cm}

Semi-supervised learning (SSL) utilizes unlabeled data to improve model performance and has achieved promising results on standard SSL image classification benchmarks~\cite{rasmus2015semi,laine2016temporal,meanteacher,mixmatch,fixmatch,noisy-student}. A common assumption, which is often made implicitly during the construction of SSL benchmark datasets, is that the class distribution of labeled and/or unlabeled data are balanced. However, in many realistic scenarios, this assumption holds untrue and becomes the primary cause of poor SSL performance~\cite{covid, darp}.

\begin{figure}
    \subfigure[]{
      \includegraphics[width=0.44\linewidth]{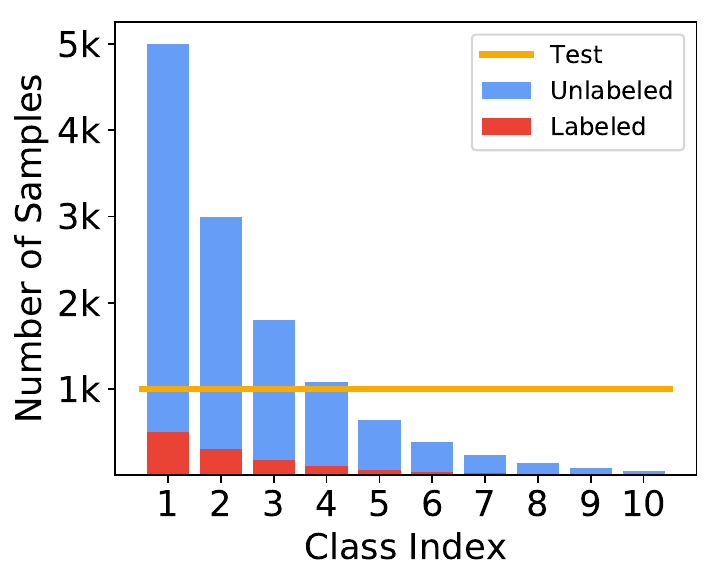}
        \label{fig:teaser_number}
    }
    \subfigure[]{
        \includegraphics[width=0.45\linewidth]{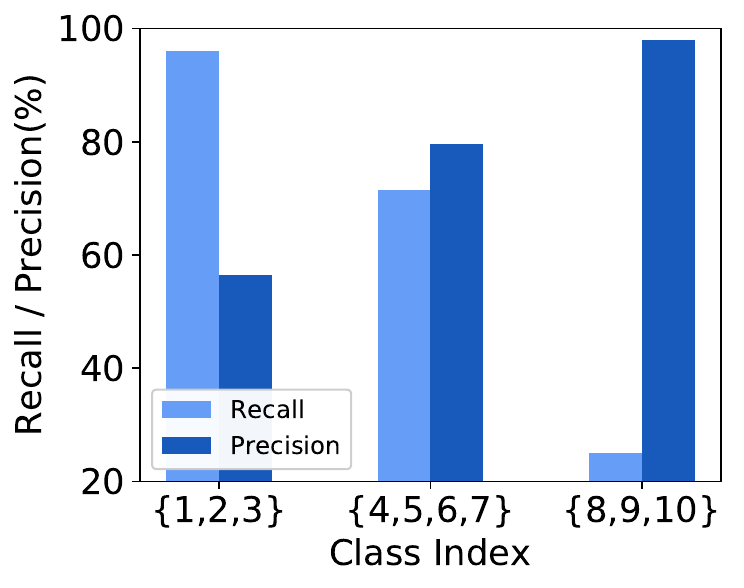}
        \label{fig:teaser_precision_recall}
    }
    \vspace{-0.3cm}
    
    \subfigure[]{
        \includegraphics[width=0.45\linewidth]{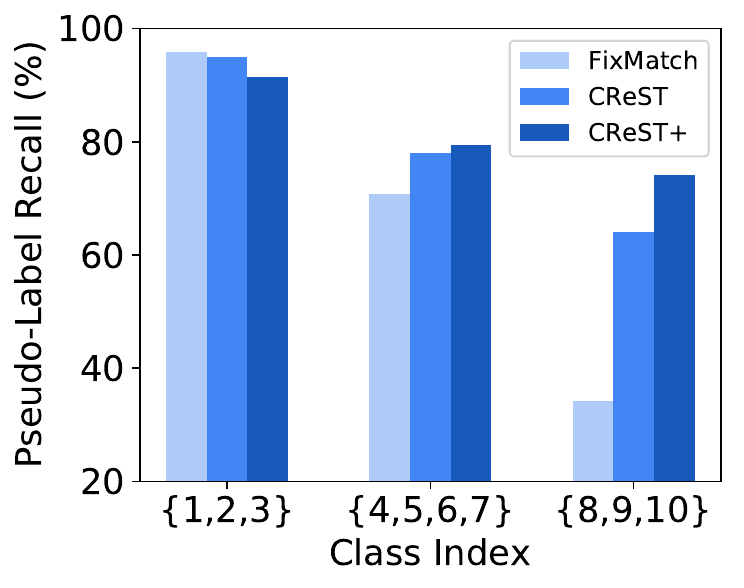}
        \label{fig:teaser_pl_acc}
    }
    \subfigure[]{
        \includegraphics[width=0.45\linewidth]{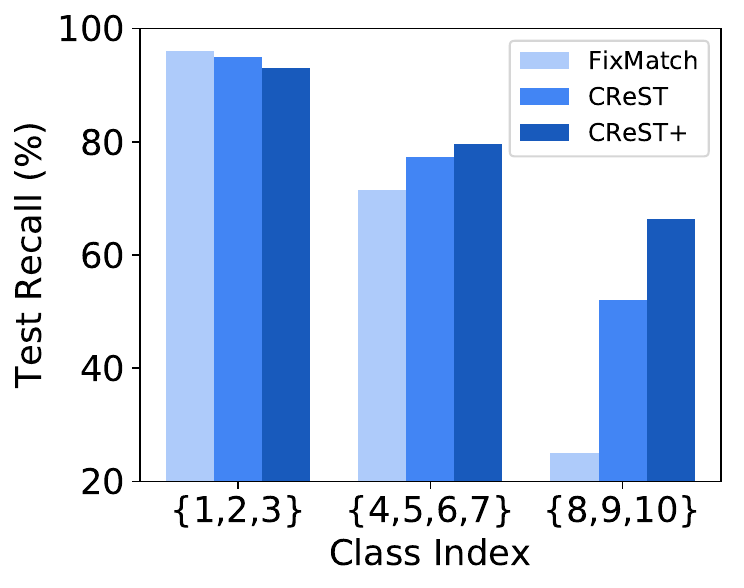}
        \label{fig:teaser_test_acc}
    }
    \caption{Experimental results on CIFAR10-LT. \subref{fig:teaser_number} Both labeled and unlabeled sets are class-imbalanced, where the most majority class has 100$\times$ more samples than the most minority class. The test set remains balanced. \subref{fig:teaser_precision_recall} Precision and recall of a FixMatch~\cite{fixmatch} model. Although minority classes have low recall, they obtain high precision. \subref{fig:teaser_pl_acc} \& \subref{fig:teaser_test_acc} The proposed CReST and CReST+ improve the quality of pseudo-labels \subref{fig:teaser_pl_acc} and thus the recall on the balanced test set \subref{fig:teaser_test_acc}, especially on minority classes.}
    \label{fig:intro}
    \vspace{-0.35cm}
\end{figure}

Supervised learning on imbalanced data has been widely explored. It is commonly observed that models trained on imbalanced data are biased towards \textit{majority classes} which have numerous examples, and away from \textit{minority classes} which have few examples. Various solutions have been proposed to help alleviate bias, such as re-sampling~\cite{rs-1,rs-2}, re-weighting~\cite{cbloss,ldam}, and two-stage training~\cite{decoupling,bbn}. All these methods rely on labels to re-balance the biased model.

In contrast, SSL on imbalanced data has been under-studied. 
In fact, data imbalance poses further challenges in SSL where missing label information precludes rebalancing the unlabeled set. Pseudo-labels for unlabeled data generated by a model trained on labeled data are commonly leveraged in SSL algorithms. However, pseudo-labels can be problematic if they are generated by an initial model trained on imbalanced data and biased toward majority classes: subsequent training with such biased pseudo-labels intensifies the bias and deteriorates the model quality. Apart from a few recent works~\cite{darp,rethinking}, the majority of existing SSL algorithms~\cite{mixmatch,remixmatch,uda,fixmatch} have not been thoroughly evaluated on imbalanced data distributions. 

In this work, we investigate SSL in the context of class-imbalanced data in which both labeled and unlabeled sets have roughly the same imbalanced class distributions, as illustrated in Fig.~\ref{fig:teaser_number}. 
We observe that the undesired performance of existing SSL algorithms on imbalanced data is mainly due to low recall on minority classes. Our method is motivated by the further observation that, despite this, precision on minority classes is surprisingly high.
In Fig.~\ref{fig:teaser_precision_recall}, we show predictions on a CIFAR10-LT dataset produced by FixMatch~\cite{fixmatch}, a representative SSL algorithm with state-of-the-art performance on balanced benchmarks. 
The model obtains high recall on majority classes but suffers from low recall on minority classes, which results in low accuracy overall on the balanced test set. However, the model has almost perfect precision on minority classes, suggesting that the model is conservative in classifying samples into minority classes, but once it makes such a prediction we can be confident it is correct. Similar observations are made on other SSL methods, and on supervised learning~\cite{jamal2020rethinking}.

With this in mind, we introduce a class-rebalancing self-training scheme (CReST) which re-trains a baseline SSL model after adaptively sampling pseudo-labeled data from the unlabeled set to supplement the original labeled set. We refer to each fully-trained baseline model as a \textit{generation}. After each generation, pseudo-labeled samples from the unlabeled set are added into the labeled set to retrain an SSL model. Rather than updating the labeled set with all pseudo-labeled samples, we instead use a \textit{stochastic} update strategy in which samples are selected with higher probability if they are predicted as minority classes, as those are more likely to be correct predictions. The updating probability is a function of the data distribution estimated from the labeled set. In addition, we extend CReST to CReST+ by incorporating distribution alignment~\cite{remixmatch} with a temperature scaling factor to control its alignment strength over generations, so that predicted data distributions are more aggressively adjusted to alleviate model bias. As shown in Fig.~\ref{fig:teaser_pl_acc} and \ref{fig:teaser_test_acc}, the proposed strategy reduces the bias of pseudo-labeling and improves the class-balanced test set accuracy as a result.

We show in experiments that CReST and CReST+ improve over baseline SSL methods by a large margin. On CIFAR-LT~\cite{cbloss,ldam}, our method outperforms FixMatch~\cite{fixmatch} under different imbalance ratios and label fractions by as much as 11.8\% in accuracy. Our method also outperforms DARP~\cite{darp}, a state-of-the-art SSL algorithm designed for learning from imbalanced data, on both MixMatch~\cite{mixmatch} and FixMatch~\cite{fixmatch} by up to 4.0\% in accuracy. To further test the efficacy of the proposed method on large-scale data, we apply our method on ImageNet127~\cite{mergedIN}, a naturally imbalanced dataset created from ImageNet~\cite{in} by merging classes based on the semantic hierarchy, and get 7.9\% gain on recall. Extensive ablation study further demonstrates that our method particularly helps improve recall on minority classes, making it a viable solution for imbalanced SSL.

%-------------------------------------------------------------------------
\section{Related work}
\subsection{Semi-supervised learning}
Recent years have observed a significant advancement of SSL research~\cite{pseudolabeling,laine2016temporal,vat,mixmatch,remixmatch,noisy-student,uda,fixmatch}. Many of these methods share similar basic techniques, such as entropy minimization~\cite{grandvalet2005semi}, pseudo-labeling, or consistency regularization, with deep learning.
Pseudo-labeling~\cite{pseudolabeling,fixmatch} trains a classifier with unlabeled data using pseudo-labeled targets derived from the model's own predictions. Relatedly, \cite{laine2016temporal,mixmatch,uda,remixmatch,noisy-student} use a model's predictive probability with temperature scaling as a soft pseudo-label. 
Consistency regularization~\cite{sajjadi2016regularization,laine2016temporal,vat} learns a classifier by promoting consistency in predictions between different views of unlabeled data, either via soft~\cite{laine2016temporal,vat,mixmatch,uda} or hard~\cite{fixmatch} pseudo-labels. 
Effective methods of generating multiple views include input data augmentations of varying strength~\cite{devries2017improved,cubuk2020randaugment,remixmatch}, standard dropout within network layers~\cite{srivastava2014dropout}, and stochastic depth~\cite{huang2016deep}.
The performance of most recent SSL methods relies on the quality of pseudo-labels. However, none of aforementioned works have studied SSL in the class-imbalanced setting, in which the quality of pseudo-labels is significantly threatened by model bias.
\subsection{Class-imbalanced supervised learning}
Research on class-imbalanced supervised learning has attracted increasing attention. Prominent works include re-sampling~\cite{smote,rs-1,rs-2,rs-3} and re-weighting~\cite{khan2017cost,cbloss,ldam,tan2020equalization}
which re-balance the contribution of each class, while others focus on re-weighting each instance~\cite{focalloss,metanet,l2rw,jamal2020rethinking}. Some works~\cite{feature-transfer,wang2019dynamic,m2m,learnable-embed,oltr} aim to transfer knowledge from majority classes to minority classes. A recent trend of work proposes to decouple the learning of representation and classifier~\cite{bbn,decoupling,tang2020causaleffect}. These methods assume all labels are available during training and their performance is largely unknown under SSL scenarios.

\begin{figure*}
    \centering
    \subfigure{
        \includegraphics[width=0.23\textwidth]{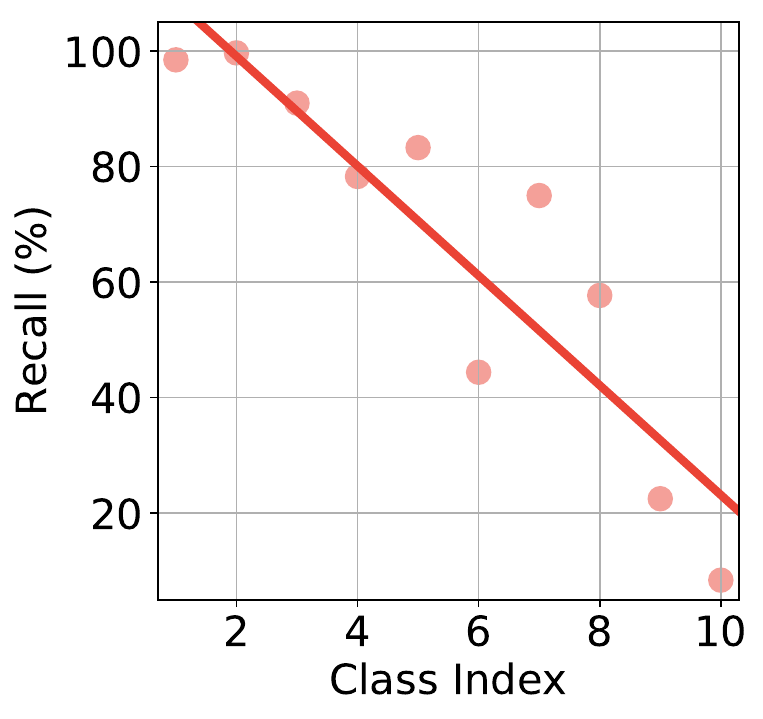}
        \label{fig:eval-a}
    }
    \centering
    \subfigure{
        \includegraphics[width=0.23\textwidth]{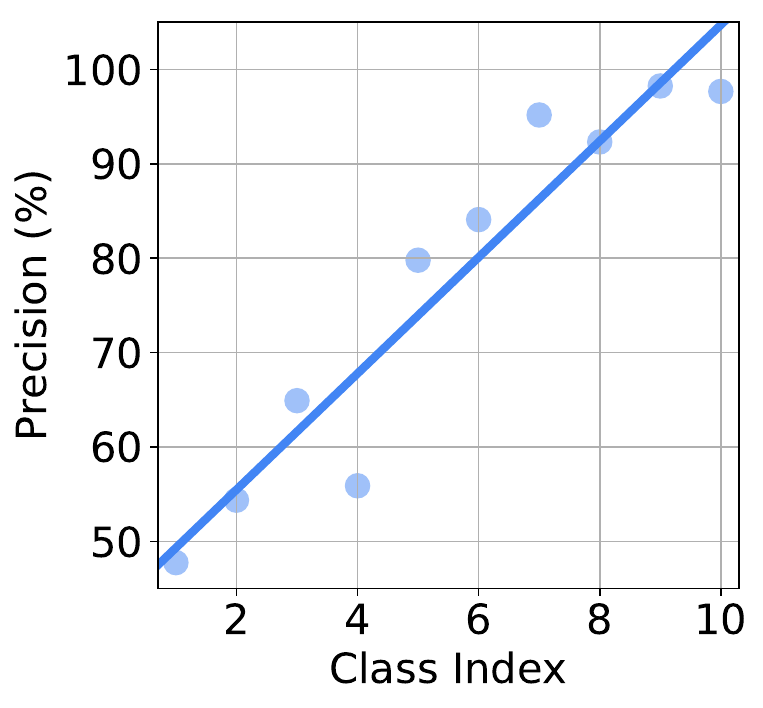}
        \label{fig:eval-b}
    }
    \centering
    \subfigure{
        \includegraphics[width=0.23\textwidth]{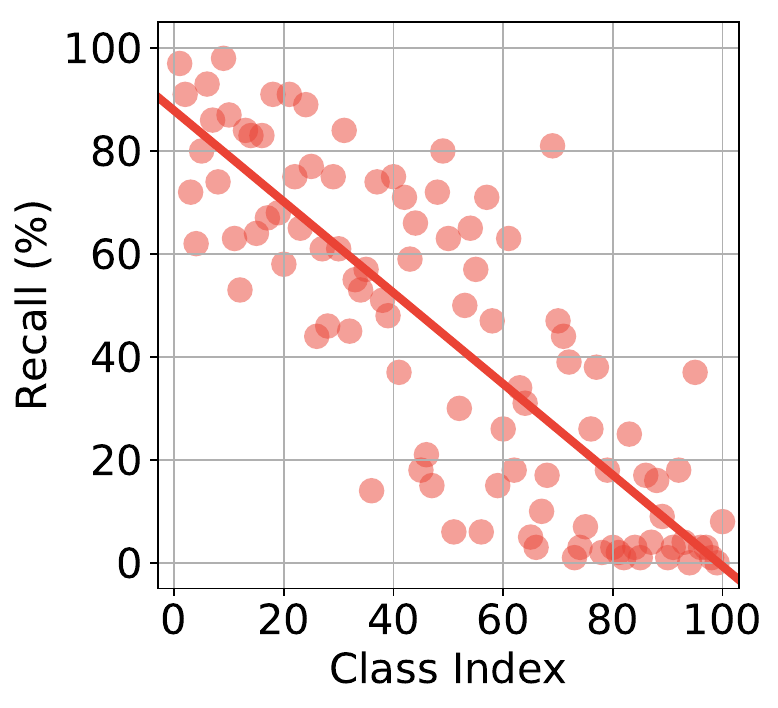}
        \label{fig:eval-c}
    }
    \centering
    \subfigure{
         \includegraphics[width=0.23\textwidth]{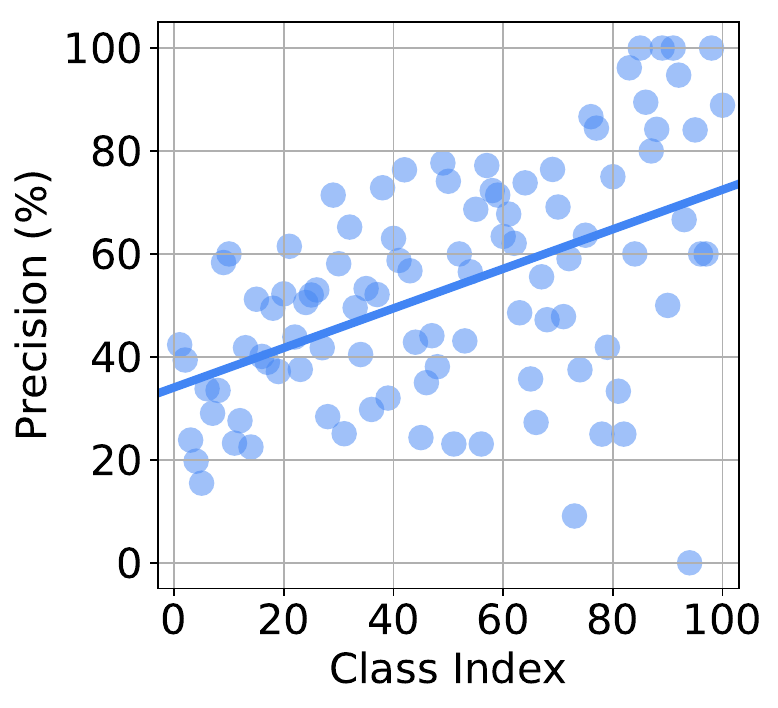}
        \label{fig:eval-d}
    }
    \caption{Bias of a FixMatch~\cite{fixmatch} model on class-imbalanced data. \textbf{Left}: Per-class recall and precision on CIFAR10-LT. \textbf{Right}: Per-class recall and precision on CIFAR100-LT. The class index is sorted by the number of examples in descending order. While the conventional assumption might be that the performance of the majority classes is better than that of the minority classes, we find it only partially true. The model obtains high recall but low precision on majority classes, while obtaining low recall but high precision on minority classes. See more details in Sec.~\ref{sec:model-bias}.}
    \label{fig:method-pa}
    \vspace{-0.3cm}
\end{figure*}

\subsection{Class-imbalanced semi-supervised learning}
While SSL has been extensively studied, it is under-explored regarding class-imbalanced data.
Recently, Yang and Xu~\cite{rethinking} argued that leveraging unlabeled
data by SSL and self-supervised learning can benefit class-imbalanced learning.
Hyun \etal~\cite{suppress} proposed a suppressed consistency loss to suppress the loss on minority classes. 
Kim \etal~\cite{darp} proposed Distribution Aligning Refinery (DARP) to refine raw pseudo-labels via a convex optimization. In contrast, we boost the quality of the model's raw pseudo-labels directly via an class-rebalancing sampling strategy and a progressive distribution alignment strategy. DARP also discussed another interesting setting where labeled and unlabeled data do not share the same class distribution, while in this work we focus on the scenario when labeled and unlabeled data have roughly the same distribution.

\section{Class-Imbalanced SSL}
\vspace{-0.1cm}
In this section, we first set up the problem and introduce baseline SSL algorithms. Next, we investigate the biased behavior of existing SSL algorithms on class-imbalanced data.
Based on these observations, we propose a class-rebalancing self-training framework (CReST) that takes advantage of, rather than suffers from, the model's bias to alleviate the performance degeneration on minority classes. 
In addition, we extend distribution alignment~\cite{remixmatch} and integrate it as CReST+ to further improve the quality of online pseudo-labeling.
\vspace{-0.1cm}
\subsection{Problem setup and baselines}
\vspace{-0.1cm}
We first set up the problem of class-imbalanced semi-supervised learning. For an $L$-class classification task, there is a labeled set $\mathcal{X}\,{=}\,\big\{(x_n, y_n) \,{:}\, n \,{\in}\, (1, \ldots, N)\big\}$, where $x_n \,{\in}\, \mathbb{R}^d$ are training examples and $y_n \,{\in}\, \{1, \ldots, L\}$ are corresponding class labels. 
The number of training examples in $\mathcal{X}$ of class $l$ is denoted as $N_l$, \ie, $\sum_{l=1}^L N_l\,{=}\,N$. 
Without loss of generality, we assume that the classes are sorted by cardinality in descending order, \ie, $N_1 \,{\geq}\, N_2 \,{\geq}\, \cdots \,{\geq}\, N_L$.
The marginal class distribution of $\mathcal{X}$ is skewed, \ie, $N_1 \,{\gg}\, N_L$. We measure the degree of class imbalance by imbalance ratio, $\gamma\,{=}\,\frac{N_1}{N_L}$. 
Besides the labeled set $\mathcal{X}$, an unlabeled set $\mathcal{U}\,{=}\,\big\{u_m \,{\in}\, \mathbb{R}^d\,{:}\, m \,{\in}\, (1, \ldots, M) \big\}$ that shares the same class distribution as $\mathcal{X}$ is also provided. The label fraction $\beta\,{=}\,\frac{N}{N+M}$ measures the percentage of labeled data. 
Given class-imbalanced sets $\mathcal{X}$ and $\mathcal{U}$, our goal is to learn a classifier $f\,{:}\, \mathbb{R}^d \,{\rightarrow}\,\{1, \ldots, L\}$ that generalizes well under a class-\textit{balanced} test criterion.

Many state-of-the-art SSL methods~\cite{fixmatch,noisy-student} utilize unlabeled data by assigning a pseudo-label with the classifier's prediction $\hat{y}_m\,{=}\,f(u_m)$.
The classifier is then optimized on both labeled and unlabeled samples with their corresponding pseudo-labels. Therefore, the quality of pseudo-labels is crucial to the final performance. 
These algorithms work successfully on standard class-balanced datasets since the quality of the classifier — and thus its online pseudo-labels — improves for all classes over the course of training. 
However, when the classifier is biased at the beginning due to a skewed class distribution, the online pseudo-labels of unlabeled data can be even more biased, further aggravating the class-imbalance issue and resulting in severe performance degradation on minority classes. 

\vspace{-0.1cm}
\subsection{A closer look at the model bias}
\label{sec:model-bias}
\vspace{-0.1cm}
Previous works~\cite{cbloss,ldam} introduce long-tailed versions of CIFAR~\cite{cifar} datasets with various class-imbalanced ratios to evaluate class-imbalanced fully-supervised learning algorithms. 
We extend this protocol by retaining a fraction of training samples as labeled and the remaining as unlabeled. 
We test FixMatch~\cite{fixmatch}, one of the state-of-the-art SSL algorithms designed for class-balanced data. Fig.~\ref{fig:method-pa} shows test recall and precision of each class on CIFAR10-LT with imbalance ratio $\gamma\,{=}\,100$, label fraction $\beta\,{=}\,10\%$, and CIFAR100-LT with imbalance ratio $\gamma\,{=}\,50$, label fraction $\beta\,{=}\,30\%$.

First, as shown in the first and third plots of Fig.~\ref{fig:method-pa}, FixMatch achieves very high recall on majority classes and poor recall on minority classes, which is consistent with the conventional wisdom. For example, the recall of the most and second most majority classes of CIFAR10-LT is 98.5\% and 99.7\%, respectively, while the model recognizes only 8.4\% of samples correctly from the most minority class. 
In other words, the model is highly biased towards majority classes, resulting in poor recall averaged over all classes which is also known as accuracy as the test set is balanced.

Despite the low recall, the minority classes maintain surprisingly high precision as in the second and fourth plots of Fig.~\ref{fig:method-pa}. For example, the model achieves 97.7\% and 98.3\% precision, respectively, on the most and the second most minority classes of CIFAR10-LT, while only achieving relatively low precision on majority classes.
This indicates that many minority class samples are predicted as one of the majority classes.

\begin{figure}
    \centering
    \includegraphics[width=0.8\linewidth]{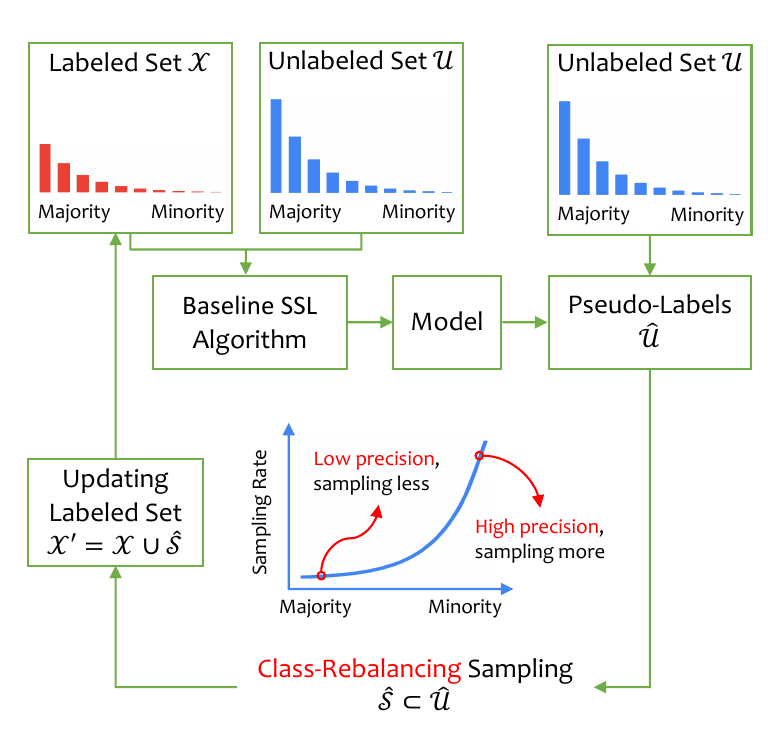}
    \caption{
    CReST (Class-Rebalancing Self-Training) alternatingly trains a baseline SSL algorithm on both labeled and unlabeled data and expands the labeled set by sampling pseudo-labeled unlabeled data. Sampling rates for majority and minority classes are adaptively determined based on the quality of pseudo-labels. See text for details.}
    \label{fig:flowchart}
    \vspace{-0.3cm}
\end{figure}

While the conventional wisdom may suggest that the performance of the majority classes is better than that of the minority classes, we find that it is only partly true: the biased model learned on class-imbalanced data indeed performs favorably on majority classes in terms of recall, but favors minority classes in terms of precision. Similar observations are made on other SSL algorithms, and also on fully-supervised class-imbalanced learning~\cite{jamal2020rethinking}. This empirical finding motivates us to exploit the high precision of minority classes to alleviate their recall degradation. To achieve this goal, we introduce CReST, a class-rebalancing self-training framework illustrated in Fig.~\ref{fig:flowchart}.

\vspace{-0.1cm}
\subsection{Class-rebalancing self-training}
\vspace{-0.1cm}
Self-training~\cite{st-1, st-2} is an iterative method widely used in SSL. 
It trains the model for multiple generations, where each generation involves two steps.
First, the model is trained on the labeled set to obtain a teacher model. 
Second, the teacher model's predictions are used to generate pseudo-labels $\hat{y}_m$ for unlabeled data $u_m$. The pseudo-labeled set $\hat{\mathcal{U}}\,{=}\,\big\{(u_m, \hat{y}_m)\big\}_{m=1}^M$ is included into the labeled set, \ie, $\mathcal{X}^\prime \,{=}\, \mathcal{X} \,{\cup}\, \hat{\mathcal{U}}$, for the next generation.

To accommodate the class-imbalance, we propose two modifications to the self-training strategy. First, instead of solely training on the labeled data, we use SSL algorithms to exploit both labeled and unlabeled data to get a better teacher model in the first step. More importantly, in the second step, rather than including every sample in $\hat{\mathcal{U}}$ in the labeled set, we instead expand the labeled set with a selected subset $\hat{\mathcal{S}} \,{\subset}\, \hat{\mathcal{U}}$, \ie, $\mathcal{X}^\prime \,{=}\, \mathcal{X} \,{\cup}\,  \hat{\mathcal{S}}$. 
We choose $\hat{\mathcal{S}}$ following a class-rebalancing rule: the less frequent a class $l$ is, the more unlabeled samples that are predicted as class $l$ are included into the pseudo-labeled set $\hat{\mathcal{S}}$.

We estimate the class distribution from the labeled set. Specifically, unlabeled samples that are predicted as class $l$ are included into $\hat{\mathcal{S}}$ at the rate of
\begin{equation}
\mu_l = \big(\frac{N_{L+1-l}}{N_1}\big)^\alpha\,,
\end{equation}
where $\alpha\,{\geq}\,0 $ tunes the sampling rate and thus the size of $\hat{\mathcal{S}}$. For instance, for a 10-class imbalanced dataset with imbalance ratio of $\gamma\,{=}\,\frac{N_1}{N_{10}}\,{=}\,100$, we keep all samples predicted as the most minority class since $\mu_{10} \,{=}\, (\frac{N_{10+1-10}}{N_1})^\alpha\,{=}\,1$. While for the most majority class, $\mu_{1}\,{=}\,(\frac{N_{10+1-1}}{N_1})^\alpha \,{=}\, 0.01^\alpha$ of samples are selected. When $\alpha \,{=}\, 0$, $\mu_l \,{=}\, 1$ for all $l$, then all unlabeled samples are kept and the algorithm falls back to the conventional self-training. When selecting pseudo-labeled samples in each class, we take the most confident ones.

The motivation of our CReST strategy is two-fold. First, as observed in Sec.~\ref{sec:model-bias}, the precision of minority classes is much higher than that of majority classes, hence minority class pseudo-labels are less risky to include in the labeled set. Second, adding samples to minority classes is more critical due to data scarcity. With more samples from minority classes added, the labeled set is more class-balanced, which leads to a less biased classifier for online pseudo-labeling in the subsequent generation. Note that there are other ways of sampling the pseudo-labels in a class-balancing fashion and we provide a practical and effective example. 

\begin{table*}[ht]
    \begin{center}
    \resizebox{\textwidth}{!}{%
    \begin{tabular}{lcccccccccc}
    \toprule
        &  \multicolumn{6}{c}{CIFAR10-LT} & \multicolumn{4}{c}{CIFAR100-LT}  \\
        \cmidrule(l{3pt}r{3pt}){2-7} \cmidrule(l{3pt}r{3pt}){8-11}
        &  \multicolumn{3}{c}{$\beta\,{=}\,10\%$} & \multicolumn{3}{c}{$\beta\,{=}\,30\%$} & \multicolumn{2}{c}{$\beta\,{=}\,10\%$}  & \multicolumn{2}{c}{$\beta\,{=}\,30\%$}  \\
        \cmidrule(l{3pt}r{3pt}){2-4} \cmidrule(l{3pt}r{3pt}){5-7} \cmidrule(l{3pt}r{3pt}){8-9} \cmidrule(l{3pt}r{3pt}){10-11}
        Method     & $\gamma\,{=}\,50$ & $\gamma\,{=}\,100$ & $\gamma\,{=}\,200$ & $\gamma\,{=}\,50$ & $\gamma\,{=}\,100$ & $\gamma\,{=}\,200$ & $\gamma\,{=}\,50$ & $\gamma\,{=}\,100$ & $\gamma\,{=}\,50$ & $\gamma\,{=}\,100$ \\
        \cmidrule(l{3pt}r{3pt}){1-1} \cmidrule(l{3pt}r{3pt}){2-4} \cmidrule(l{3pt}r{3pt}){5-7} \cmidrule(l{3pt}r{3pt}){8-9} \cmidrule(l{3pt}r{3pt}){10-11}
        FixMatch~\cite{fixmatch}   & 79.4\ms{0.65} & 66.3\ms{1.74} & 59.7\ms{0.74} & 81.9\ms{0.30} & 73.1\ms{0.58} & 64.7\ms{0.69} & 33.7\ms{0.94} & 28.3\ms{0.66} & 43.1\ms{0.24} & 38.6\ms{0.45}\\
        %FixMatch & \checkmark & \\
        w/ CReST   & 83.8\ms{0.45} & 75.9\ms{0.62} & 64.1\ms{0.23} & 84.2\ms{0.13} & 77.6\ms{0.86} & 67.7\ms{0.82} & 37.4\ms{0.29} & 32.1\ms{1.52} & 45.6\ms{0.19} & 40.2\ms{0.53}\\
        w/ CReST+  & \textbf{84.2}\ms{0.39} & \textbf{78.1}\ms{0.84} & \textbf{67.7}\ms{1.39} & \textbf{84.9}\ms{0.27} & \textbf{79.2}\ms{0.20} & \textbf{70.5}\ms{0.56} & \textbf{38.8}\ms{1.03} & \textbf{34.6}\ms{0.74} & \textbf{46.7}\ms{0.34} & \textbf{42.0}\ms{0.44}\\
        \bottomrule
    \end{tabular}
    }%
    \end{center}
    \vspace{-0.1cm}
    \caption{Classification accuracy (\%) on CIFAR10-LT and CIFAR100-LT under various label fraction $\beta$ and imbalance ratio $\gamma$. The numbers are averaged over 5 different folds. Models with CReST are trained for 15 generations. Models with CReST+ are trained for 6 generations.}
    \label{tab:cifar-main-results}
    \vspace{-0.3cm}
\end{table*}

\vspace{-0.1cm}
\subsection{Progressive distribution alignment}
\label{sec:adaptive-da}
\vspace{-0.1cm}
We further improve the quality of online pseudo-labels by additionally introducing progressive distribution alignment into CReST and distinguish it as CReST+.

While first introduced for class-balanced SSL, Distribution Alignment (DA)~\cite{remixmatch} fits with class-imbalanced scenarios particularly well. It aligns the model's predictive distribution on unlabeled samples with the labeled training set's class distribution $p(y)$. Let $\tilde{p}(y)$ be the moving average of the model’s predictions on unlabeled examples. DA first scales the model’s prediction $q\,{=}\,p(y | u_m;f)$ for an unlabeled example $u_m$ by the ratio $\frac{p(y)}{\tilde{p}(y)}$, aligning $q$ with the target distribution $p(y)$. It then re-normalizes the scaled result to form a valid probability distribution: $\tilde{q}\,{=}\,\text{Normalize}(q \frac{p(y)}{\tilde{p}(y)})$, where $\text{Normalize}(x)_i\,{=}\,x_i / \sum_j x_j$. $\tilde{q}$ is used as the label guess for $u_m$ instead of $q$. 

To further enhance DA's ability to handle class-imbalanced data, we extend it with temperature scaling. Specifically, we add a tuning knob $t\,{\in}\, [0,1]$ that controls the class-rebalancing strength of DA. Instead of directly taking $p(y)$ as target, we use a temperature-scaled distribution $\text{Normalize}(p(y)^t)$. When $t\,{=}\,1$, we recover DA. When $t\,{<}\,1$, the temperature-scaled distribution becomes smoother and balances the model’s predictive distribution more aggressively. When $t\,{=}\,0$, the target distribution becomes uniform.

While using a smaller $t$ can benefit a single generation under a class-balanced test criterion, it is less desirable for multiple generations of self-training since it affects the quality of pseudo-labels.
Specifically, applying a $t\,{<}\,1$ enforces the model's predictive distribution to be more balanced than the class distribution of the training set, leading the model to predict minority classes more frequently. 
However, on an imbalanced training set with few samples of minority classes, such pseudo-labeling tends to be over-balanced, \ie, more samples are wrongly predicted as minority classes. This decreases the high precision of minority classes, interfering with our ability to exploit it to produce better pseudo-labels.

To handle this, we propose to progressively increase the strength of class-rebalancing by decreasing $t$ over generations. Specifically, we set $t$ by a linear function of the current generation $g$ which indexes from 0: 
\begin{equation}
t_g = (1-\frac{g}{G})\cdot 1.0  + \frac{g}{G} \cdot t_{\text{min}}\,,
\end{equation}
where $G\,{+}\,1$ is the total number of generations and $t_{\text{min}}$ is the temperature used for the last generation. This progressive schedule for $t$ enjoys both high precision of pseudo-labels in early generations, and stronger class-rebalancing in late generations. It also speeds up the iterative training, obtaining better results with fewer generations of training. See Sec.~\ref{sec:ablation} for empirical analysis.

\vspace{-0.1cm}
\section{Experiments}
\vspace{-0.1cm}
\subsection{CIFAR-LT}
\vspace{-0.05cm}
\paragraph{Datasets.} We first evaluate the efficacy of the proposed method on long-tailed CIFAR10 (CIFAR10-LT) and long-tailed CIFAR100 (CIFAR100-LT) introduced in \cite{cbloss,ldam}. On these datasets, training images are randomly discarded per class to maintain a pre-defined imbalance ratio $\gamma$. Specifically, $N_l\,{=}\,\gamma^{-\frac{l-1}{L-1}}\,{\cdot}\, N_1$ while $N_1\,{=}\,5000$, $L\,{=}\,10$ for CIFAR10-LT and $N_1\,{=}\,500$, $L\,{=}\,100$ for CIFAR100-LT. We randomly select $\beta\,{=}\,10$\% and $30$\% of samples from training data to create the labeled set, and test imbalance ratio $\gamma\,{=}\,50$, $100$ and $200$ for CIFAR10-LT and $\gamma\,{=}\,50$ and $100$ for CIFAR100-LT. The test set remains untouched and balanced, so that the evaluated criterion, accuracy on the test set, is class-balanced.

\vspace{-0.05cm}
\paragraph{Setup.} We use Wide ResNet-28-2~\cite{wrn} following \cite{realistic, fixmatch} as the backbone. We evaluate our method on FixMatch and MixMatch. For each generation, the model is trained for $2^{16}$ steps when using FixMatch as the baseline SSL algorithm and $2^{17}$ steps for MixMatch. We use a cosine learning rate decay~\cite{cosinelr,fixmatch} whose formulation is provided in the supplementary material. Other hyper-parameters for each training generation are untouched. For CReST and CReST+ related hyper-parameters, we set $\alpha\,{=}\,1\,{/}\,3$, $t_{\text{min}}\,{=}\,0.5$ for FixMatch and $\alpha\,{=}\,1\,{/}\,2$, $t_{\text{min}}\,{=}\,0.8$ for MixMatch. CReST takes 15 generations, while CReST+ only takes 6 generations accelerated by progressive distribution alignment.
The hyper-parameters are selected based on a single fold of CIFAR10-LT with $\gamma\,{=}\,100$ and $\beta\,{=}\,10\%$.
We evaluate the model on the test dataset every $2^{10}$ steps and report the average test accuracy of the last 5 evaluations.
Each algorithm is tested under 5 different folds of labeled data and we report the mean and the standard deviation of accuracy on the test set. Following \cite{mixmatch} and \cite{fixmatch}, we report final performance using an exponential moving average of model parameters.

\vspace{-0.05cm}
\paragraph{Main results.} First, we compare our model with baseline FixMatch, and present the results in Table~\ref{tab:cifar-main-results}. Although FixMatch performs reasonably well on imbalance ratio $\gamma\,{=}\,50$, its accuracy decreases significantly with increasing imbalance ratio. 
In contrast, CReST improves the accuracy of FixMatch on all evaluated settings and achieves as much as 9.6\% absolute performance gain. When incorporating progressive distribution alignment, our CReST+ model is able to further boost the performance on all settings by another few points, resulting in 3.0\% to 11.8\% absolute accuracy improvement compared to baseline FixMatch.

The accuracy of all compared methods improves with increasing number of labeled samples, but CReST consistently outperforms the baseline. 
This indicates that CReST can better utilize labeled data to reduce model bias under imbalanced class-distribution.

We also observe that our model works particularly well and achieves 11.8\% and 6.1\% accuracy gain for imbalance ratio $\gamma\,{=}\,100$ with $10\%$ and $30\%$ labeled data, respectively. We hypothesize the reason is that our model finds more correctly pseudo-labeled samples to augment the labeled set when the imbalance ratio is moderate. However, when imbalance ratio is very high, \eg, $\gamma\,{=}\,200$, our model's capability is constrained by insufficient number of training samples from minority classes. 

\vspace{-0.05cm}
\paragraph{Comparison with baselines.} We further report the performance of other SSL baselines in Table~\ref{tab:cifar-other-baselines}. For fair comparison, all algorithms are trained for $6\,{\times}\,2^{16}$ steps. This leads to 6 generations for CReST and CReST+ on a FixMatch base with $2^{16}$ steps each generation, and 3 generations for CReST and CReST+ on a MixMatch base with $2^{17}$ steps each generation. Other models that do not use self-training are trained for a single generation with $6\,{\times}\,2^{16}$ steps.

\begin{table}[ht]
    \begin{center}
    \begin{tabular}{lccc}
    \toprule
        Method          &    $\gamma\,{=}\,50$   & $\gamma\,{=}\,100$  & $\gamma\,{=}\,200$ \\
         \cmidrule(l{3pt}r{3pt}){1-1}  \cmidrule(l{3pt}r{3pt}){2-4}
        Pseudo-Labeling~\cite{pseudolabeling} &    52.5\ms{0.74} & 46.5\ms{1.29} & 42.0\ms{1.39} \\
        Mean Teacher~\cite{meanteacher}    &    57.1\ms{3.00} & 48.1\ms{0.71} & 45.1\ms{1.28} \\
        \cmidrule(l{3pt}r{3pt}){1-1}  \cmidrule(l{3pt}r{3pt}){2-4}
        MixMatch~\cite{mixmatch}        &    69.1\ms{1.18} & 60.4\ms{2.24} & 54.5\ms{1.87} \\
        w/ CReST        &    69.8\ms{1.06} & 60.5\ms{1.56} & 55.2\ms{2.25} \\
        w/ CReST+       &    76.7\ms{0.35} & 66.1\ms{0.79} & 57.6\ms{1.30} \\
         \cmidrule(l{3pt}r{3pt}){1-1}  \cmidrule(l{3pt}r{3pt}){2-4}
        FixMatch~\cite{fixmatch}        &    80.1\ms{0.44} & 67.3\ms{1.19} & 59.7\ms{0.63} \\
        w/ CB~\cite{cbloss}           &    80.2\ms{0.45} & 67.6\ms{1.88} & 60.8\ms{0.26} \\
        w/ RS~\cite{rs-1,rs-2}         &    80.2\ms{0.78} & 69.6\ms{1.30} & 60.9\ms{1.25} \\
        w/ DA~\cite{remixmatch} ($t\,{=}\,1.0$)           &    80.2\ms{0.45} & 69.7\ms{1.27} & 62.0\ms{0.84} \\
        w/ DA~\cite{remixmatch} ($t\,{=}\,0.5$)           &    82.4\ms{0.33} & 73.6\ms{0.63} & 63.7\ms{1.17} \\
        w/ LA~\cite{la}           &    83.2\ms{0.87} & 70.4\ms{2.90} & 62.4\ms{1.24} \\
        w/ CReST        &    83.2\ms{0.37} & 74.8\ms{1.09} & 63.4\ms{0.32} \\
        w/ CReST+       &    84.2\ms{0.39} & 78.1\ms{0.84} & 67.7\ms{1.39} \\
        w/ CReST+ \& LA &    \textbf{85.6}\ms{0.36} & \textbf{81.2}\ms{0.70} & \textbf{71.9}\ms{2.24}  \\
        \bottomrule
    \end{tabular}
    \end{center}
    \vspace{-0.1cm}
    \caption{We compare CReST and CReST+ with baseline methods including different SSL algorithms and typical class-rebalancing techniques designed for fully-supervised learning. For fair comparison, all models are measured at the same number of training steps. See text for details. Three imbalance ratios $\gamma$ with $\beta\,{=}\,10\%$ labels are evaluated. Numbers are averaged over 5 different folds.} 
    \label{tab:cifar-other-baselines}
    \vspace{-0.3cm}
\end{table}

\begin{table}[ht]
    \begin{center}
    \begin{tabular}{lccc}
    \toprule
        Method       &    $\gamma\,{=}\,50$ & $\gamma\,{=}\,100$ & $\gamma\,{=}\,150$ \\
        \cmidrule(l{3pt}r{3pt}){1-1}  \cmidrule(l{3pt}r{3pt}){2-4}
        Supervised & 65.2\ms{0.05} & 58.8\ms{0.13} & 55.6\ms{0.43}  \\
        \cmidrule(l{3pt}r{3pt}){1-1}  \cmidrule(l{3pt}r{3pt}){2-4}
        MixMatch~\cite{mixmatch}   & 73.2\ms{0.56} & 64.8\ms{0.28} & 62.5\ms{0.31}  \\
        w/ DARP~\cite{darp}    & 75.2\ms{0.47} & 67.9\ms{0.14} & 65.8\ms{0.52}  \\
        w/ CReST   & 78.4\ms{0.36} & 70.0\ms{0.49} & 64.7\ms{0.96}  \\
        w/ CReST+  & \textbf{79.0}\ms{0.26} & \textbf{71.9}\ms{0.33} & \textbf{68.3}\ms{0.57}  \\
        \cmidrule(l{3pt}r{3pt}){1-1}  \cmidrule(l{3pt}r{3pt}){2-4}
        FixMatch~\cite{fixmatch}   & 79.2\ms{0.33} & 71.5\ms{0.72} & 68.4\ms{0.15}  \\
        w/ DARP~\cite{darp}    & 81.8\ms{0.24} & 75.5\ms{0.05} & 70.4\ms{0.25}  \\
        w/ CReST   & 83.0\ms{0.39} & 75.7\ms{0.38} & 70.8\ms{0.25}  \\
        w/ CReST+  & \textbf{83.9}\ms{0.14} & \textbf{77.4}\ms{0.36} & \textbf{72.8}\ms{0.58}  \\
        \bottomrule
    \end{tabular}
    \end{center}
    \vspace{-0.1cm}
    \caption{Accuracy (\%) under DARP's protocol~\cite{darp} on CIFAR10. See the supplementary material for dataset details. Three imbalance ratios $\gamma$ are evaluated. Numbers are averaged over 5 runs.} 
    \label{tab:cifar-darp}
\end{table}

\begin{table}[ht]
    \begin{center}
    \begin{tabular}{lccc}
    \toprule
        Method                    & Gen$_1$ & Gen$_2$ & Gen$_3$ \\
         \cmidrule(l{3pt}r{3pt}){1-1}  \cmidrule(l{3pt}r{3pt}){2-4}
        Supervised (100\% labels) & 75.8 & -     & -  \\
        Supervised (10\% labels)  & 46.0 & -     & -  \\
        FixMatch (10\% labels)    & 65.8 & -     & -  \\
        w/ DA ($t\,{=}\,0.5$)     & 69.1 & -     & -  \\
        w/ CReST                  & 65.8 & 67.6 & 67.7 \\
        w/ CReST+                 & 68.3 & 70.7 & \textbf{73.7} \\
        \bottomrule
    \end{tabular}
    \end{center}
    \vspace{-0.1cm}
    \caption{Evaluating the proposed method on ImageNet127 with $\beta\,{=}\,10\%$ samples are labeled. We retrain FixMatch models for 3 generations with our CReST and CReST+.} 
    \label{tab:imagenet127}
    \vspace{-0.25cm}
\end{table}

We first directly evaluate several classic SSL methods on class-imbalanced datasets, including Pseudo-Labeling~\cite{pseudolabeling}, Mean Teacher~\cite{meanteacher}, MixMatch~\cite{mixmatch} and FixMatch~\cite{fixmatch}. 
All the SSL baselines suffer from low accuracy due to imbalanced data, and the accuracy drop becomes more pronounced with increasing imbalance ratio.
On MixMatch, the improvement provided by CReST is modest mainly due to the schedule constraint. Providing more generation budget, the results of MixMatch with CReST can be further improved.
Among these algorithms, FixMatch achieves the best performance, so we take it as the baseline for various rebalancing methods. 

We consider typical class-rebalancing methods designed for fully-supervised learning that can be directly applied in SSL algorithms including 1) Class-Balanced loss (CB)~\cite{cbloss}, a representative of re-weighting strategies in which labeled examples are re-weighted according to the inverse of the effective number of samples in each class; 2) Re-Sampling (RS)~\cite{rs-1, rs-2}, a representative of re-sampling strategies in which each labeled example is sampled with probability proportional to the inverse sample size of its class. We also consider Distribution Alignment (DA)~\cite{remixmatch} as described in Sec.~\ref{sec:adaptive-da} and Logit Adjustment (LA)~\cite{la}, an ad-hoc post-processing technique to enhance models' discriminative ability on minority classes by adjusting the logits of model predictions. 
While CB, RS, DA and LA all improve accuracy over SSL baselines, the gain is relatively small. With CReST and CReST+, we successfully improve the accuracy for all imbalance ratios by at most 10.8\% over FixMatch, outperforming all compared SSL baselines and class-rebalancing methods. 
Finally, applying LA as the post-processing correction of our CReST+ models further gives consistent accuracy gains, producing the best results.

\vspace{-0.06cm}
\paragraph{Comparison with DARP.} We directly compare with DARP~\cite{darp}, the most recent state-of-the-art SSL algorithm specifically designed for imbalanced data. Both DARP and our method are built upon MixMatch and FixMatch as drop-in additions to standard SSL algorithms. We apply our method on exactly the same datasets used in DARP and present the results in Table~\ref{tab:cifar-darp}. Details of the dataset construction are provided in the supplementary material. For all three imbalance ratios, our model consistently achieves up to 4.0\% accuracy gain over DARP on MixMatch, and up to 2.4\% accuracy gain on FixMatch. 

\subsection{ImageNet127}
\vspace{-0.06cm}
\paragraph{Datasets.} We also evaluate CReST on ImageNet127~\cite{mergedIN, nca} to verify its performance on large-scale datasets. 
ImageNet127 is originally introduced in \cite{mergedIN}, where the 1000 classes of ImageNet~\cite{in} are grouped into 127 classes based on their top-down hierarchy in WordNet. It is a naturally imbalanced dataset with imbalance ratio $\gamma\,{=}\,286$. Its most majority class ``mammal'' consists of 218 original classes and 277,601 training images. While its most minority class ``butterfly'' is formed by a single original class with 969 training examples. We randomly select $\beta\,{=}\,10$\% training samples as the labeled set and keep the test set unchanged. Due to class grouping, the test set is not balanced. Therefore, we compute averaged class recall instead of accuracy to achieve a balanced metric. 

We note that there are other large-scale datasets like iNaturalist~\cite{inaturalists} and ImageNet-LT~\cite{oltr} which often serve as testbeds for fully-supervised long-tailed recognition algorithms. However, these datasets contain too few examples of minority classes to form a statistically meaningful dataset and draw reliable conclusions for semi-supervised learning. For example, there are only 5 examples in the most minority class of the ImageNet-LT dataset. 

\vspace{-0.06cm}
\paragraph{Setup.} We use ResNet50~\cite{resnet} as the backbone. 
The hyper-parameters for each training generation are adopted from the original FixMatch paper. 
The model is self-trained for 3 generations with $\alpha\,{=}\,0.7$ and $t_{\text{min}}\,{=}\,0.5$.

\vspace{-0.06cm}
\paragraph{Results.} We report results in Table~\ref{tab:imagenet127}. Supervised learning with 100\% and 10\% labeled training examples and DA with temperature scaling are also presented as reference. Comparing with the baseline FixMatch, both CReST and CReST+ progressively improve over 3 generations of self-training, while CReST+ provides 7.9\% absolute gain in the end, which verifies the efficacy of our proposed method.

\begin{figure}
    \subfigure[]{
      \includegraphics[width=0.415\linewidth]{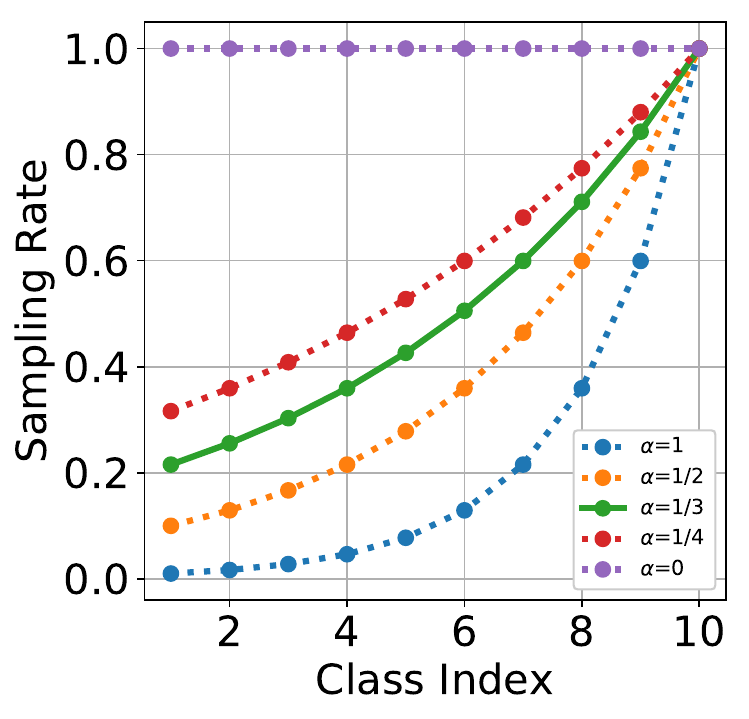}
      \label{fig:ablation-alpha-a}
    }
    \subfigure[]{
      \includegraphics[width=0.545\linewidth]{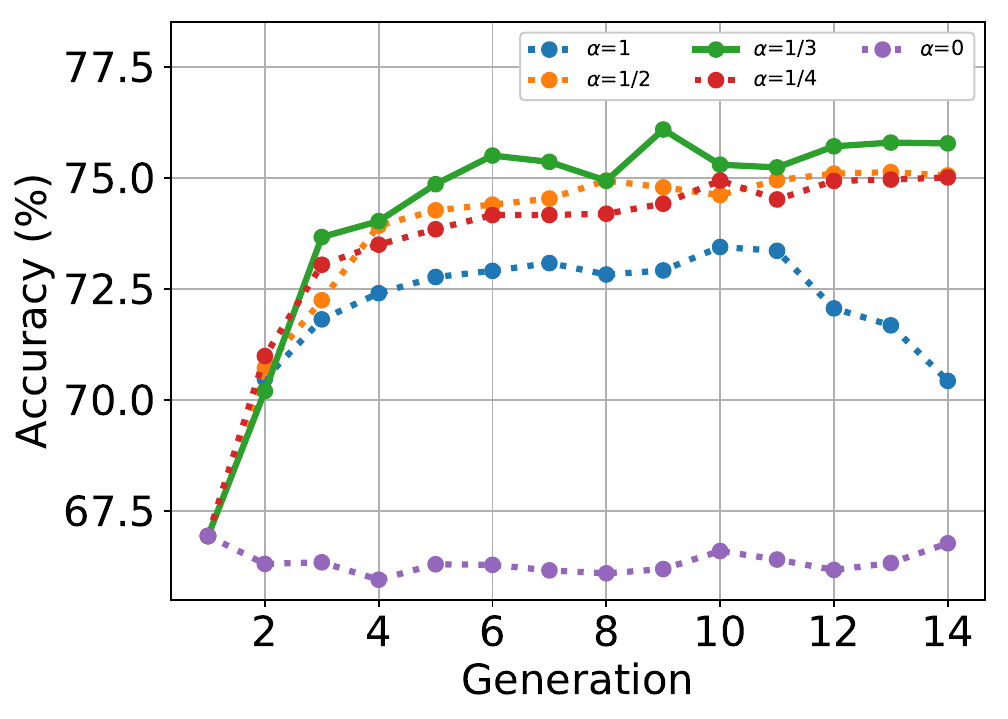}
      \label{fig:ablation-alpha-b}
    }
    \caption{Effect of $\alpha$ across multiple generations on CIFAR10-LT ($\gamma\,{=}\,100$, $\beta\,{=}\,10\%$) in CReST. \subref{fig:ablation-alpha-a} Illustration of how $\alpha$ influences sampling rate. \subref{fig:ablation-alpha-b} Test accuracy over generations with different $\alpha$. When $\alpha\,{=}\,0$, the method falls back to conventional self-training with all the unlabeled examples and corresponding pseudo-labels added into the labeled set, showing no improvement after generations of retraining, whereas our class-rebalancing sampling ($\alpha\,{>}\,0$) helps.}
    \label{fig:ablation-alpha}
    \vspace{-0.2cm}
\end{figure}

\begin{figure}
    \subfigure[]{
      \includegraphics[width=0.425\linewidth]{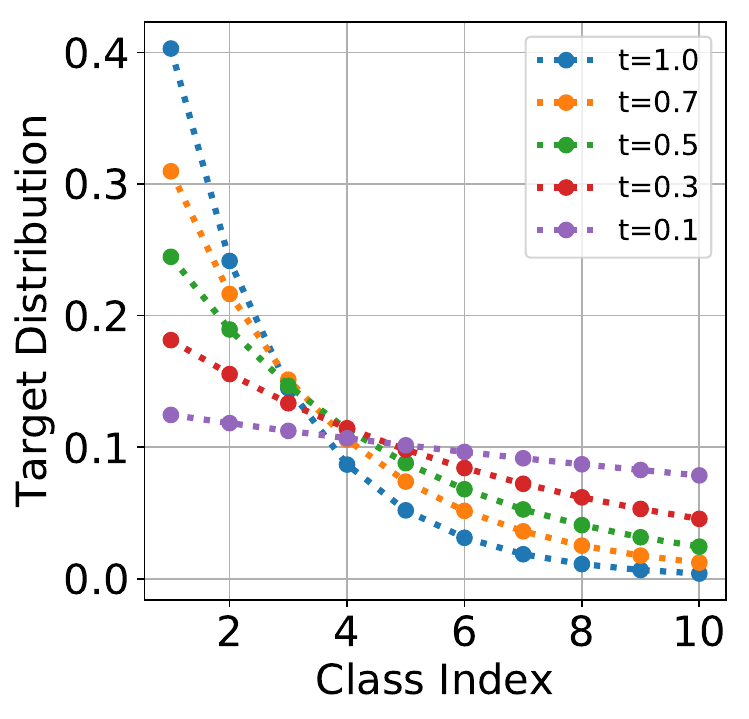}
      \label{fig:ablation-t-a}
    }
    \subfigure[]{
      \includegraphics[width=0.54\linewidth]{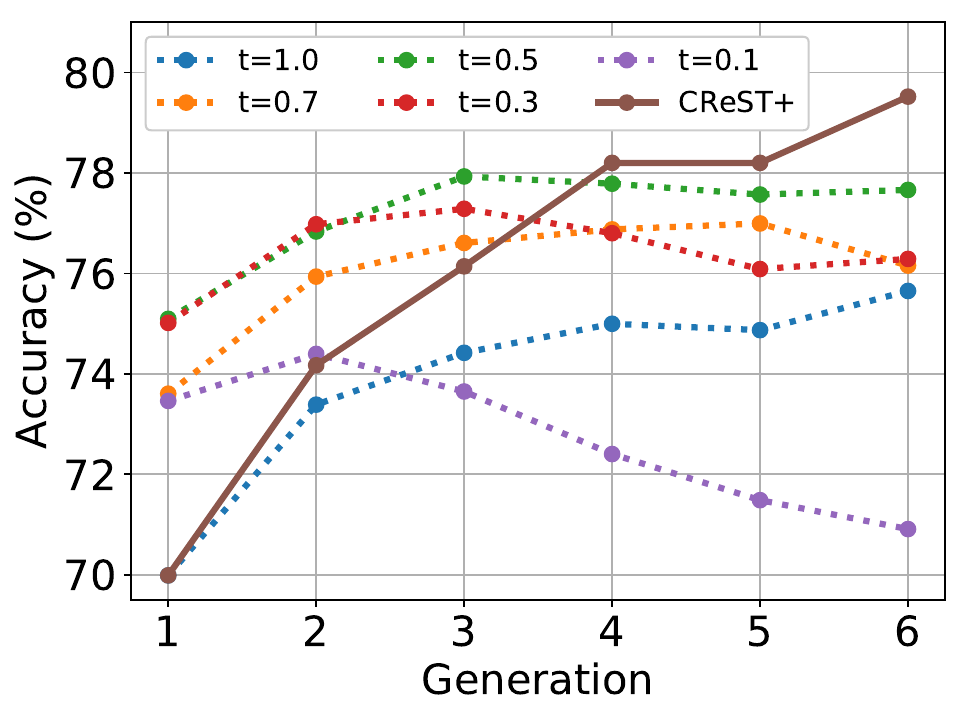}
      \label{fig:ablation-t-b}
    }
    \caption{Effect of temperature $t$ across multiple generations on CIFAR10-LT ($\gamma\,{=}\,100$, $\beta\,{=}\,10\%$). \subref{fig:ablation-t-a} Illustration of how $t$ controls the target distribution of distribution alignment. \subref{fig:ablation-t-b} Test accuracy over generations with different constant $t$ and our CReST+ using progressive $t$. Compared to using a constant $t$, CReST+ achieves the best final accuracy by progressing from $t\,{=}\,0$ to $t_{\text{min}}\,{=}\,0.5$ over 6 generations.} 
    \label{fig:ablation-t}
    \vspace{-0.2cm}
\end{figure}

\begin{table*}[ht]
    \begin{center}
    \begin{tabular}{lcrrrrrrrrrrr}
    \toprule
        Method /\ Class & Split          & 1    & 2    & 3    & 4    & 5    & 6    & 7    & 8    & 9   & 10 & Avg. \\
        \cmidrule(l{3pt}r{3pt}){1-1} \cmidrule(l{3pt}r{3pt}){2-2} \cmidrule(l{3pt}r{3pt}){3-12} \cmidrule(l{3pt}r{3pt}){13-13}
        FixMatch~\cite{fixmatch}        & test           &\textbf{98.7} & \textbf{99.5} & \textbf{90.0} & \textbf{83.5} & 85.0 & 47.6 & 69.9 & 59.0 & 8.9  & 7.2  & 64.9 \\
        w/ CReST        & test           &97.7 & 98.3 & 88.8 & 81.9 & \textbf{88.2} & 59.7 & 79.5 & 61.2 & 47.0 & 47.9 & 75.0 \\
\rowcolor{Gray}         &                &-1.0 & -1.2 & -1.2 & -1.6 & +3.2 & +12.1& +9.6 & +2.2 & +38.1& +40.7& +10.1\\
        w/ CReST+       & test           &93.8 & 97.7 & 87.3 & 76.9 & 87.5 & \textbf{69.2} & \textbf{84.9} & \textbf{67.9} & \textbf{60.3} & \textbf{70.8} & \textbf{79.6} \\
\rowcolor{Gray}         &                &-4.9 & -1.8 & -2.7 & -6.6 & +2.5 & +21.6& +15.0& +8.9 & +51.4& +63.6& +14.7 \\
        \cmidrule(l{3pt}r{3pt}){1-1} \cmidrule(l{3pt}r{3pt}){2-2} \cmidrule(l{3pt}r{3pt}){3-12} \cmidrule(l{3pt}r{3pt}){13-13}
        FixMatch~\cite{fixmatch}        & unlabeled      &\textbf{98.5} & \textbf{99.1} & \textbf{90.0} & \textbf{84.0} & 84.7 & 49.7 & 64.9 & 65.6 & 14.9 & 22.2 & 67.4 \\
        w/ CReST        & unlabeled      &97.8 & 96.8 & \textbf{90.0} & 82.9 & 87.4 & 62.4 & 79.3 & 64.8 & 60.8 & 66.7 & 78.9 \\
\rowcolor{Gray}         &                &-0.7 & -2.3 & 0    & -1.1 & +2.7 & +12.7& +14.4& -0.8 & +45.9& +44.5& +11.5\\
        w/ CReST+       & unlabeled      &92.2 & 95.7 & 86.1 & 76.7 & \textbf{87.6} & \textbf{68.1} & \textbf{85.1} & \textbf{71.2} & \textbf{75.7} & \textbf{75.6} & \textbf{81.4} \\
\rowcolor{Gray}         &                &-6.3 & -3.4 & -3.9 & -7.3 & +2.9 & +18.4& +20.2& +5.6 & +60.8& +53.4& +14.0 \\
        \bottomrule
    \end{tabular}
    \end{center}
    \caption{Per-class recall (\%) on the balanced test set and the imbalanced unlabeled set of CIFAR10-LT ($\gamma\,{=}\,100$, $\beta\,{=}\,10\%$). Our strategies compromise small loss on majority classes for significant gain on minority classes, leading to improved averaged recall over all classes.}
    \label{tab:ablation-break-down}
    \vspace{-0.3cm}
\end{table*}

\subsection{Ablation study}
\label{sec:ablation}
We perform an extensive ablation study to evaluate and understand the contribution of each critical component in CReST and CReST+. The experiments in this section are all performed with FixMatch on CIFAR10-LT with imbalance ratio $\gamma\,{=}\,100$, label fraction $\beta\,{=}\,10\%$ and a single fold of labeled data.

\vspace{-0.06cm}
\paragraph{Effect of sampling rate.}
CReST introduces the sampling rate hyper-parameter $\alpha$ that controls the per-class sampling rate and the number of selected pseudo-labeled samples to be included in the labeled set. In Fig.~\ref{fig:ablation-alpha} we show how $\alpha$ influences performance over generations.

When $\alpha\,{=}\,0$, our method falls back to conventional self-training, which expands the labeled set with all unlabeled examples and their corresponding predicted labels. 
However, conventional self-training does not produce a performance gain over multiple generations, showing that simply applying self-training can not provide performance improvement. 
In contrast, with our class-rebalancing sampling strategy ($\alpha\,{>}\,0$), the accuracy can be improved by iterative model retraining.

As illustrated in Fig.~\ref{fig:ablation-alpha-a}, smaller $\alpha$ means more pseudo-labeled samples are added into the labeled set, which enlarges the labeled set but adversely introduces more low-quality pseudo-labels. 
On the other hand, larger $\alpha$ biases pseudo-labeled samples towards minority classes. 
As a result, the class-rebalancing sampling can be too strong with large $\alpha$, leading to imbalance in the reversed direction, towards the original minority classes. This is the case for $\alpha\,{=}\,1$ where, after the 10-\textit{th} generation, the model becomes increasingly biased towards minority classes and suffers from performance degradation on majority classes, resulting in decreased accuracy. For example, from the 10-\textit{th} generation to the last generation, the recall of the most minority classes increases by a large margin from 55.0\% to 71.1\%, while 7 of the other 9 classes suffer from severe recall degradation, resulting in 3.0\% drop of the class-balanced test set accuracy.
Empirically, we find $\alpha\,{=}\,1\,{/}\,3$ achieves a balance between the quality of pseudo-labels and the class-rebalancing strength across different imbalance ratios and label fractions on long-tailed CIFAR datasets.

\vspace{-0.05cm}
\paragraph{Effect of progressive temperature scaling.} The proposed adaptive distribution alignment used in CReST+ introduces another hyper-parameter, temperature $t$, that scales the target distribution.
We first illustrate in Fig.~\ref{fig:ablation-t-a} how temperature $t$ smooths the target distribution in distribution alignment so that smaller $t$ provides stronger re-balancing strength. In Fig.~\ref{fig:ablation-t-b}, we study the effect of using a constant temperature and our proposed progressive temperature scaling in which $t$ gradually decreases from $1.0$ to $t_{\min}\,{=}\,0.5$ across generations of self-training.

First, we notice that $t\,{=}\,0.5$ provides the best \textit{single} generation accuracy of 75.1\% among all tested temperature values. This suggests that the model can benefit from class re-balancing with a properly ``smoothed'' target distribution compared with 70.0\% accuracy of the original distribution alignment whose temperate $t$ is fixed to $1.0$. 
Further decreasing $t$ to $0.1$ results in lower accuracy, as the target distribution is overly smoothed, which introduces more pseudo-labeling errors.

Over multiple generations of self-training, using a constant $t$ is not optimal. Although a relatively small $t$ (\eg, 0.5) can give better performance in early generations, it can not provide further gains through continuing self-training due to the decreased pseudo-label quality. 
When $t$ is lower than 0.5, performance can even degrade after certain later generations. 
In contrast, the proposed CReST+, which progressively enhances the distribution alignment strength, provides the best accuracy at the last generation.

\paragraph{Per-class performance.} To show the source of accuracy improvements, in Table~\ref{tab:ablation-break-down} we present per-class recall on the balanced test set of CIFAR10-LT with imbalance ratio 100 and label fraction 10\%.
Both CReST and CReST+ sacrifice a few points of accuracy on four majority classes but provide significant gains on the other six minority classes, obtaining better performance over all classes.
We also include the results on the imbalanced unlabeled set. The results are particularly similar to those of the test set with mild drop on majority classes and remarkable improvement on minority classes. This suggests that our method indeed improves the quality of pseudo-labels, which can be transferred to better generalization on a balanced test criterion.

\section{Conclusion}

In this work, we present a class-rebalancing self-training framework, named CReST, for imbalanced semi-supervised learning. 
CReST is motivated by the observation that existing SSL algorithms produce high precision pseudo-labels on minority classes.  
CReST iteratively refines a baseline SSL model by supplementing the labeled set with high quality pseudo-labels, where minority classes are updated more aggressively than majority classes. Over generations of self-training, the model becomes less biased towards majority classes and focuses more on minority classes.
We also extend distribution alignment to progressively increase its class-rebalancing strength over generations and denote the combined method CReST+. 
Extensive experiments on long-tailed CIFAR datasets and ImageNet127 dataset demonstrate that the proposed CReST and CReST+ improve baseline SSL algorithms by a large margin, and consistently outperform state-of-the-art rebalancing methods.

\paragraph{Acknowledgments.} We thank Zizhao Zhang, Yin Cui and Prannay Khosla for valuable advice. We thank Alex Kurakin for helping experiments on ImageNet, Jinsung Yoon for the proofread of our manuscript. This work is partially supported by ONR N00014-20-1-2206 to CW and AY.

\appendix
\section{Appendix}

\subsection{Details on learning rate schedule}
Following FixMatch~\cite{fixmatch}, we use cosine learning rate decay~\cite{cosinelr} which sets the learning rate to $\eta\cos{(\frac{l\pi k}{16K})}$ where $l$ is a hyper-parameter controlling the decay rate, $\eta$ is the initial learning rate, $k$ is the current training step, and $K$ is the total number of training steps per generation. We use $l=1$ for CReST and CReST+ on a FixMatch~\cite{fixmatch} base, on $l=5$ for CReST models on a MixMatch~\cite{mixmatch} base.

\subsection{Details on datasets}

\paragraph{Details of DARP datasets.} We directly compare with DARP~\cite{darp}, the most recent state-of-the-art SSL algorithm specifically designed for class-imbalanced data. We apply our method on exactly the same datasets used in DARP, which are class-imbalanced datasets constructed from CIFAR10~\cite{cifar}. Similar to long-tailed CIFAR10 (CIFAR10-LT), the training images are randomly selected per class. However, instead of maintaining  different label fractions $\beta$, DAPR keeps the number of training samples in the most majority class $N_1\,{=}\,4500$, including $3000$ unlabeled samples and $1500$ labeled samples. That is, the label fraction is set to be $\beta\,{=}\,\frac{1500}{4500}=33.3\%$ This setting is applied to all evaluated imbalance ratios $\gamma\,{=}\,50, 100, 150$. Please refer to DAPR~\cite{darp} for more details.

\begin{figure}
    \centering
    \includegraphics[width=0.9\linewidth]{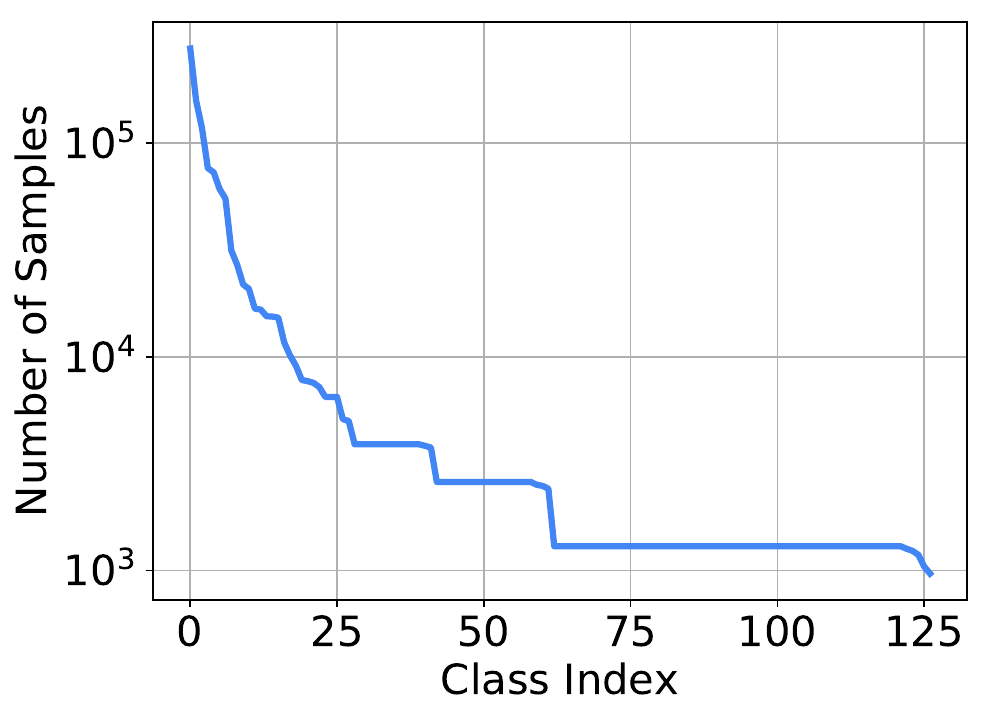}
    \caption{The imbalanced class distribution of the training set of ImageNet127~\cite{mergedIN}.}
    \label{fig:in127}
\end{figure}

\paragraph{Class distribution of ImageNet127.} We apply our method on ImageNet127~\cite{mergedIN} to test the efficacy of CReST and CReST+ on a large-scale dataset. Both the training set and the validation set of ImageNet127 are built by grouping the 1000 classes of ImageNet~\cite{in} into 127 classes based on their top-down hierarchy in WordNet. It is originally introduced to study the relationship between coarse classes and their fine-grained classes. And in \cite{mergedIN} and \cite{nca}, only the instance-wise accuracy on the imbalanced validation set is considered which is a class-imbalanced metric. In our work, however, we focus on its naturally class-imbalanced property, and our metric is class-balanced, \ie, averaged recall over all classes.

\begin{figure}
    \centering
    \subfigure[Pseudo-Labeling~\cite{pseudolabeling}]{
       \subfigure{
       \includegraphics[width=0.47\linewidth]{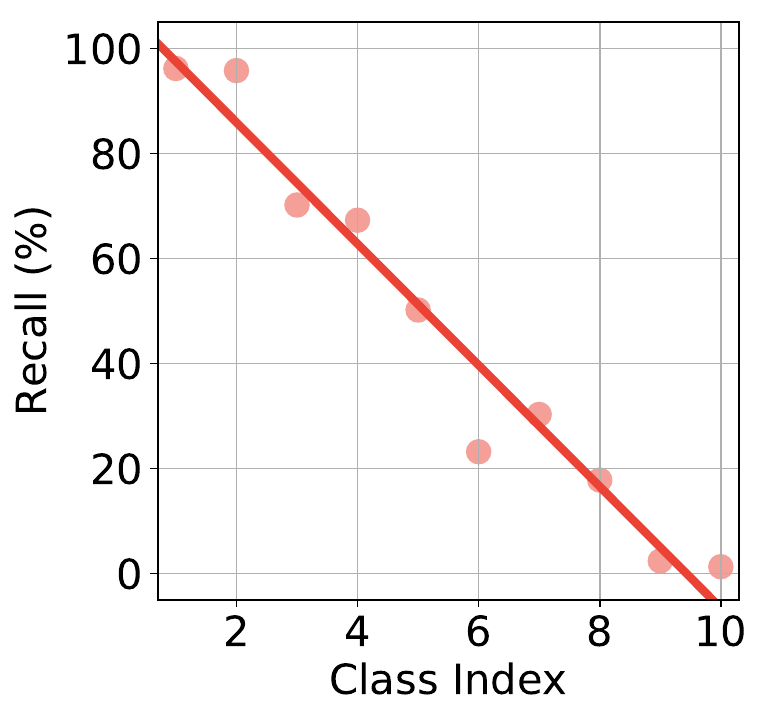}
       }
       \subfigure{
       \includegraphics[width=0.47\linewidth]{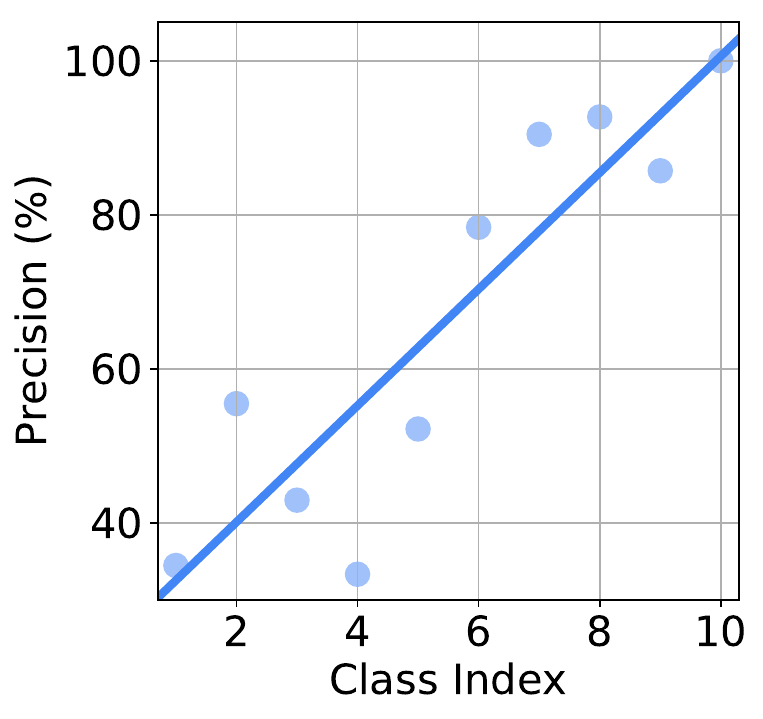}
       }
    }

    \subfigure[Mean Teacher~\cite{meanteacher}]{
        \subfigure{
        \includegraphics[width=0.47\linewidth]{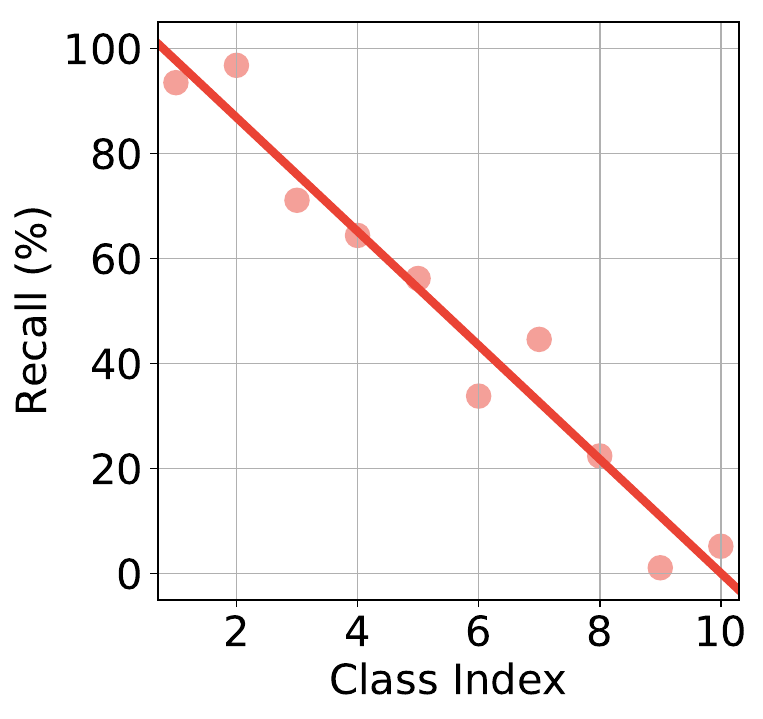}
        }
        \subfigure{
        \includegraphics[width=0.47\linewidth]{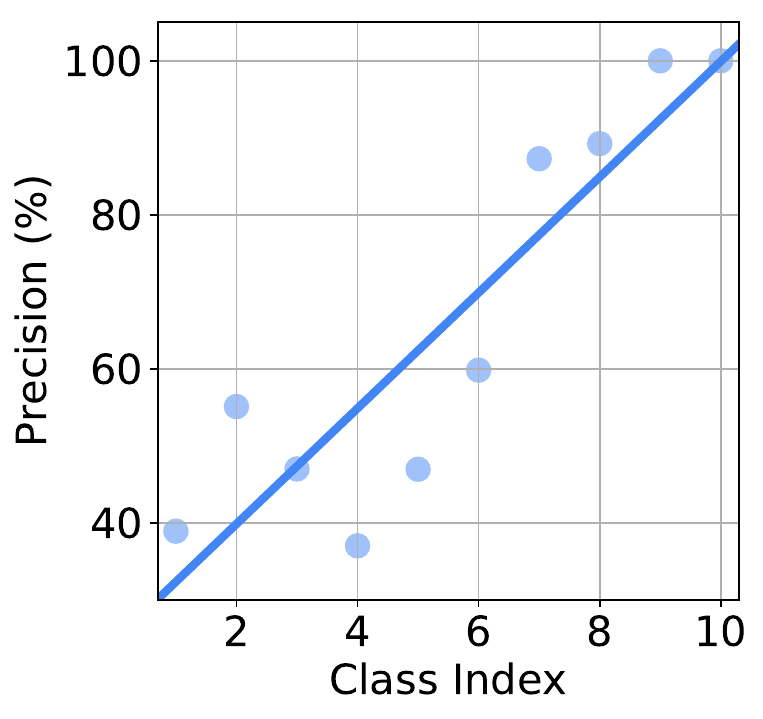}
        }
    }
    
    \subfigure[MixMatch~\cite{mixmatch}]{
        \subfigure{
        \includegraphics[width=0.47\linewidth]{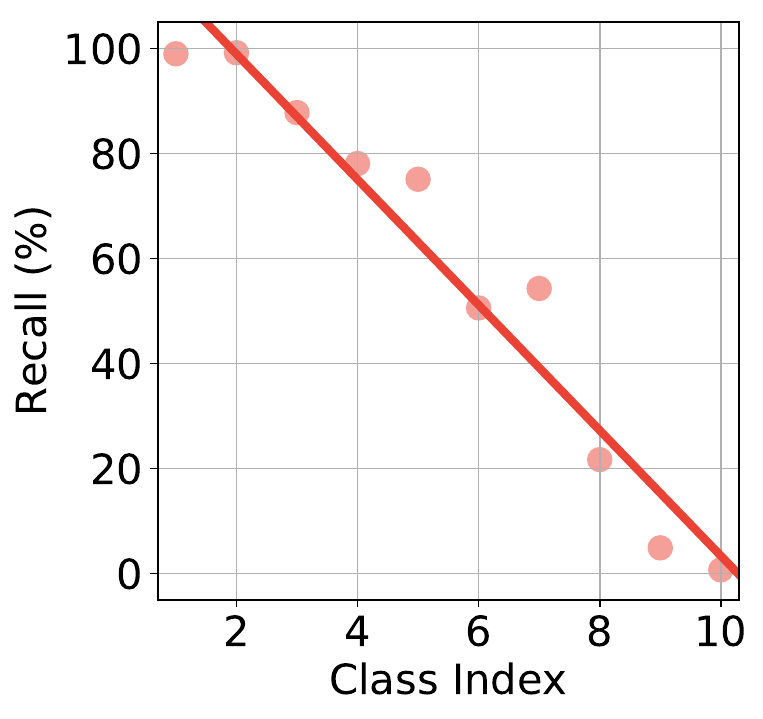}
        }
        \subfigure{
        \includegraphics[width=0.47\linewidth]{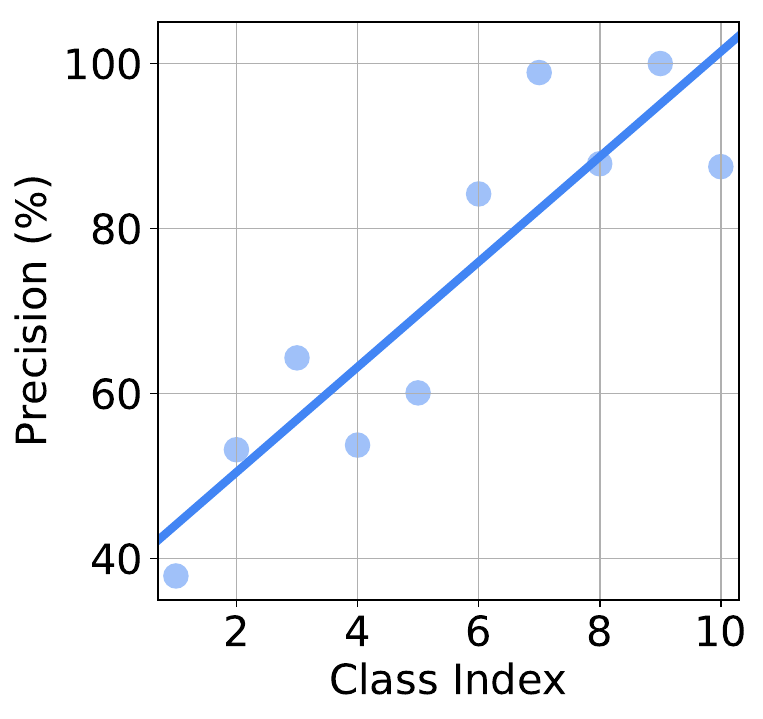}
        }
    }
    \caption{Per-class recall and precision on CIFAR10-LT ($\gamma\,{=}\,100$, $\beta\,{=}\,10\%$) with three different SSL algorithms. The class index is sorted by the number of examples in descending order. In line with our observation on FixMatch~\cite{fixmatch}, these models obtain high recall but low precision on majority classes, while obtaining low recall but high precision on minority classes.}
    \label{fig:othersll}
\end{figure}

In Fig.~\ref{fig:in127}, we show the number of samples of each merged class on the training set. The class index is sorted by the size of the class in descending order. Note that the y-axis is log-scaled. As shown in Fig.~\ref{fig:in127}, the class-distribution is highly skewed, while the most minority class still contains 969 training examples, which is adequate to be split to labeled and unlabeled subsets to form an SSL task. This is unlike common benchmark datasets designed for fully-supervised learning (\eg, ~\cite{oltr,inaturalists}) where the minority classes have too few examples to bulld an valid SSL task. These two properties, \ie the skewed class-distribution and the adequate training samples in minority classes, make ImageNet127 a good test bed for class-imblanced semi-supervised learning.

\subsection{Precision and recall of other SSL algorithms}
In this section, we provide recall and precision of each class with three different SSL algorithms, including Pseudo-Labeling~\cite{pseudolabeling}, Mean Teacher~\cite{meanteacher} and MixMatch~\cite{mixmatch}. We directly apply these three SSL algorithms on CIFAR10-LT with imbalanced ratio 100 and label fraction 10\%. The results are presented in Fig.~\ref{fig:othersll}. All three algorithms behave similarly, where majority classes obtain high recall but low precision, and minority classes suffer from low recall but achieve surprisingly high precision. The opposite bias of recall and precision of these models is in line with our observation made on FixMatch~\cite{fixmatch}. This common phenomenon shared by different SSL algorithms motivates our to exploit the high precision of minority classes to alleviate their recall degradation, re-balancing the model during the process of self-training.

{
\small
\bibliographystyle{ieee_fullname}
\bibliography{egbib}

\begin{thebibliography}{10}\itemsep=-1pt

\bibitem{remixmatch}
David Berthelot, Nicholas Carlini, Ekin~D. Cubuk, Alex Kurakin, Kihyuk Sohn,
  Han Zhang, and Colin Raffel.
\newblock {ReMixMatch}: Semi-supervised learning with distribution matching and
  augmentation anchoring.
\newblock In {\em ICLR}, 2020.

\bibitem{mixmatch}
David Berthelot, Nicholas Carlini, Ian Goodfellow, Nicolas Papernot, Avital
  Oliver, and Colin~A Raffel.
\newblock {MixMatch}: A holistic approach to semi-supervised learning.
\newblock In {\em NeurIPS}, 2019.

\bibitem{rs-1}
Mateusz Buda, Atsuto Maki, and Maciej~A Mazurowski.
\newblock A systematic study of the class imbalance problem in convolutional
  neural networks.
\newblock {\em Neural Networks}, 2018.

\bibitem{rs-2}
Jonathon Byrd and Zachary Lipton.
\newblock What is the effect of importance weighting in deep learning?
\newblock In {\em ICML}, 2019.

\bibitem{covid}
Saul Calderon-Ramirez, Armaghan Moemeni, David Elizondo, Simon
  Colreavy-Donnelly, Luis~Fernando Chavarria-Estrada, Miguel~A Molina-Cabello,
  et~al.
\newblock Correcting data imbalance for semi-supervised covid-19 detection
  using x-ray chest images.
\newblock {\em arXiv preprint arXiv:2008.08496}, 2020.

\bibitem{ldam}
Kaidi Cao, Colin Wei, Adrien Gaidon, Nikos Arechiga, and Tengyu Ma.
\newblock Learning imbalanced datasets with label-distribution-aware margin
  loss.
\newblock In {\em NeurIPS}, 2019.

\bibitem{smote}
Nitesh~V Chawla, Kevin~W Bowyer, Lawrence~O Hall, and W~Philip Kegelmeyer.
\newblock {SMOTE}: synthetic minority over-sampling technique.
\newblock {\em Journal of Artificial Intelligence Research}, 2002.

\bibitem{cubuk2020randaugment}
Ekin~D Cubuk, Barret Zoph, Jonathon Shlens, and Quoc~V Le.
\newblock Randaugment: Practical automated data augmentation with a reduced
  search space.
\newblock In {\em CVPR Workshop}, 2020.

\bibitem{cbloss}
Yin Cui, Menglin Jia, Tsung-Yi Lin, Yang Song, and Serge Belongie.
\newblock Class-balanced loss based on effective number of samples.
\newblock In {\em CVPR}, 2019.

\bibitem{inaturalists}
Yin Cui, Yang Song, Chen Sun, Andrew Howard, and Serge Belongie.
\newblock Large scale fine-grained categorization and domain-specific transfer
  learning.
\newblock In {\em CVPR}, 2018.

\bibitem{in}
Jia Deng, Wei Dong, Richard Socher, Li-Jia Li, Kai Li, and Li Fei-Fei.
\newblock {ImageNet}: A large-scale hierarchical image database.
\newblock In {\em CVPR}, 2009.

\bibitem{devries2017improved}
Terrance DeVries and Graham~W Taylor.
\newblock Improved regularization of convolutional neural networks with cutout.
\newblock {\em arXiv preprint arXiv:1708.04552}, 2017.

\bibitem{grandvalet2005semi}
Yves Grandvalet and Yoshua Bengio.
\newblock Semi-supervised learning by entropy minimization.
\newblock In {\em NeurIPS}, 2005.

\bibitem{rs-3}
Haibo He and Edwardo~A Garcia.
\newblock Learning from imbalanced data.
\newblock {\em IEEE Transactions on Knowledge and Data Engineering},
  21(9):1263--1284, 2009.

\bibitem{resnet}
Kaiming He, Xiangyu Zhang, Shaoqing Ren, and Jian Sun.
\newblock Deep residual learning for image recognition.
\newblock In {\em CVPR}, 2016.

\bibitem{huang2016deep}
Gao Huang, Yu Sun, Zhuang Liu, Daniel Sedra, and Kilian~Q Weinberger.
\newblock Deep networks with stochastic depth.
\newblock In {\em ECCV}, 2016.

\bibitem{mergedIN}
Minyoung Huh, Pulkit Agrawal, and Alexei~A Efros.
\newblock What makes imagenet good for transfer learning?
\newblock {\em arXiv preprint arXiv:1608.08614}, 2016.

\bibitem{suppress}
Minsung Hyun, Jisoo Jeong, and Nojun Kwak.
\newblock Class-imbalanced semi-supervised learning.
\newblock {\em arXiv preprint arXiv:2002.06815}, 2020.

\bibitem{jamal2020rethinking}
Muhammad~Abdullah Jamal, Matthew Brown, Ming-Hsuan Yang, Liqiang Wang, and
  Boqing Gong.
\newblock Rethinking class-balanced methods for long-tailed visual recognition
  from a domain adaptation perspective.
\newblock In {\em CVPR}, 2020.

\bibitem{decoupling}
Bingyi Kang, Saining Xie, Marcus Rohrbach, Zhicheng Yan, Albert Gordo, Jiashi
  Feng, and Yannis Kalantidis.
\newblock Decoupling representation and classifier for long-tailed recognition.
\newblock In {\em ICLR}, 2020.

\bibitem{khan2017cost}
Salman~H Khan, Munawar Hayat, Mohammed Bennamoun, Ferdous~A Sohel, and Roberto
  Togneri.
\newblock Cost-sensitive learning of deep feature representations from
  imbalanced data.
\newblock {\em IEEE transactions on neural networks and learning systems},
  2017.

\bibitem{darp}
Jaehyung Kim, Youngbum Hur, Sejun Park, Eunho Yang, Sung~Ju Hwang, and Jinwoo
  Shin.
\newblock Distribution aligning refinery of pseudo-label for imbalanced
  semi-supervised learning.
\newblock In {\em NeurIPS}, 2020.

\bibitem{m2m}
Jaehyung Kim, Jongheon Jeong, and Jinwoo Shin.
\newblock M2m: Imbalanced classification via major-to-minor translation.
\newblock In {\em CVPR}, 2020.

\bibitem{cifar}
Alex Krizhevsky and Geoffrey Hinton.
\newblock Learning multiple layers of features from tiny images.
\newblock 2009.

\bibitem{laine2016temporal}
Samuli Laine and Timo Aila.
\newblock Temporal ensembling for semi-supervised learning.
\newblock {\em arXiv preprint arXiv:1610.02242}, 2016.

\bibitem{pseudolabeling}
Dong-Hyun Lee.
\newblock Pseudo-label: The simple and efficient semi-supervised learning
  method for deep neural networks.
\newblock In {\em ICML Workshop}, 2013.

\bibitem{focalloss}
Tsung-Yi Lin, Priya Goyal, Ross Girshick, Kaiming He, and Piotr Doll{\'a}r.
\newblock Focal loss for dense object detection.
\newblock In {\em ICCV}, 2017.

\bibitem{learnable-embed}
Jialun Liu, Yifan Sun, Chuchu Han, Zhaopeng Dou, and Wenhui Li.
\newblock Deep representation learning on long-tailed data: A learnable
  embedding augmentation perspective.
\newblock In {\em CVPR}, 2020.

\bibitem{oltr}
Ziwei Liu, Zhongqi Miao, Xiaohang Zhan, Jiayun Wang, Boqing Gong, and Stella~X.
  Yu.
\newblock Large-scale long-tailed recognition in an open world.
\newblock In {\em CVPR}, 2019.

\bibitem{cosinelr}
Ilya Loshchilov and Frank Hutter.
\newblock {SGDR}: Stochastic gradient descent with warm restarts.
\newblock In {\em ICLR}, 2017.

\bibitem{la}
Aditya~Krishna Menon, Sadeep Jayasumana, Ankit~Singh Rawat, Himanshu Jain,
  Andreas Veit, and Sanjiv Kumar.
\newblock Long-tail learning via logit adjustment.
\newblock {\em arXiv preprint arXiv:2007.07314}, 2020.

\bibitem{vat}
Takeru Miyato, Shin-ichi Maeda, Masanori Koyama, and Shin Ishii.
\newblock Virtual adversarial training: a regularization method for supervised
  and semi-supervised learning.
\newblock {\em TPAMI}, 2018.

\bibitem{realistic}
Avital Oliver, Augustus Odena, Colin~A Raffel, Ekin~Dogus Cubuk, and Ian
  Goodfellow.
\newblock Realistic evaluation of deep semi-supervised learning algorithms.
\newblock In {\em NeurIPS}, 2018.

\bibitem{rasmus2015semi}
Antti Rasmus, Mathias Berglund, Mikko Honkala, Harri Valpola, and Tapani Raiko.
\newblock Semi-supervised learning with ladder networks.
\newblock In {\em Advances in neural information processing systems}, pages
  3546--3554, 2015.

\bibitem{l2rw}
Mengye Ren, Wenyuan Zeng, Bin Yang, and Raquel Urtasun.
\newblock Learning to reweight examples for robust deep learning.
\newblock In {\em ICML}, 2018.

\bibitem{sajjadi2016regularization}
Mehdi Sajjadi, Mehran Javanmardi, and Tolga Tasdizen.
\newblock Regularization with stochastic transformations and perturbations for
  deep semi-supervised learning.
\newblock {\em NeurIPS}, 2016.

\bibitem{st-1}
H Scudder.
\newblock Probability of error of some adaptive pattern-recognition machines.
\newblock {\em IEEE Transactions on Information Theory}, 1965.

\bibitem{metanet}
Jun Shu, Qi Xie, Lixuan Yi, Qian Zhao, Sanping Zhou, Zongben Xu, and Deyu Meng.
\newblock Meta-weight-net: Learning an explicit mapping for sample weighting.
\newblock In {\em NeurIPS}, 2019.

\bibitem{fixmatch}
Kihyuk Sohn, David Berthelot, Chun-Liang Li, Zizhao Zhang, Nicholas Carlini,
  Ekin~D Cubuk, Alex Kurakin, Han Zhang, and Colin Raffel.
\newblock {FixMatch}: Simplifying semi-supervised learning with consistency and
  confidence.
\newblock In {\em NeurIPS}, 2020.

\bibitem{srivastava2014dropout}
Nitish Srivastava, Geoffrey Hinton, Alex Krizhevsky, Ilya Sutskever, and Ruslan
  Salakhutdinov.
\newblock Dropout: a simple way to prevent neural networks from overfitting.
\newblock {\em JMLR}, 2014.

\bibitem{tan2020equalization}
Jingru Tan, Changbao Wang, Buyu Li, Quanquan Li, Wanli Ouyang, Changqing Yin,
  and Junjie Yan.
\newblock Equalization loss for long-tailed object recognition.
\newblock In {\em CVPR}, 2020.

\bibitem{tang2020causaleffect}
Kaihua Tang, Jianqiang Huang, and Hanwang Zhang.
\newblock Long-tailed classification by keeping the good and removing the bad
  momentum causal effect.
\newblock In {\em NeurIPS}, 2020.

\bibitem{meanteacher}
Antti Tarvainen and Harri Valpola.
\newblock Mean teachers are better role models: Weight-averaged consistency
  targets improve semi-supervised deep learning results.
\newblock In {\em NeurIPS}, 2017.

\bibitem{wang2019dynamic}
Yiru Wang, Weihao Gan, Jie Yang, Wei Wu, and Junjie Yan.
\newblock Dynamic curriculum learning for imbalanced data classification.
\newblock In {\em ICCV}, 2019.

\bibitem{nca}
Zhirong Wu, Alexei~A Efros, and Stella~X Yu.
\newblock Improving generalization via scalable neighborhood component
  analysis.
\newblock In {\em ECCV}, 2018.

\bibitem{uda}
Qizhe Xie, Zihang Dai, Eduard Hovy, Minh-Thang Luong, and Quoc~V. Le.
\newblock Unsupervised data augmentation for consistency training.
\newblock In {\em NeurIPS}, 2020.

\bibitem{noisy-student}
Qizhe Xie, Minh-Thang Luong, Eduard Hovy, and Quoc~V Le.
\newblock Self-training with noisy student improves imagenet classification.
\newblock In {\em CVPR}, 2020.

\bibitem{rethinking}
Yuzhe Yang and Zhi Xu.
\newblock Rethinking the value of labels for improving class-imbalanced
  learning.
\newblock In {\em NeurIPS}, 2017.

\bibitem{st-2}
David Yarowsky.
\newblock Unsupervised word sense disambiguation rivaling supervised methods.
\newblock In {\em 33rd annual meeting of the association for computational
  linguistics}, 1995.

\bibitem{feature-transfer}
Xi Yin, Xiang Yu, Kihyuk Sohn, Xiaoming Liu, and Manmohan Chandraker.
\newblock Feature transfer learning for deep face recognition with
  under-represented data.
\newblock In {\em CVPR}, 2018.

\bibitem{wrn}
Sergey Zagoruyko and Nikos Komodakis.
\newblock Wide residual networks.
\newblock In {\em BMVC}, 2016.

\bibitem{bbn}
Boyan Zhou, Quan Cui, Xiu-Shen Wei, and Zhao-Min Chen.
\newblock Bbn: Bilateral-branch network with cumulative learning for
  long-tailed visual recognition.
\newblock In {\em CVPR}, 2020.

\end{thebibliography}
}

\end{document}